\documentclass[10pt,twocolumn,letterpaper]{article}

\usepackage{wacv}              

\newcommand{\sd}[1]{\,{\tiny(#1)}}

\definecolor{wacvblue}{rgb}{0.21,0.49,0.74}
\usepackage[pagebackref,breaklinks,colorlinks,allcolors=wacvblue]{hyperref}
\usepackage{amsmath, amssymb, algorithm, algpseudocode, multirow}
\usepackage{xcolor}
\usepackage{cuted}
\usepackage{capt-of}
\usepackage{soul}
\usepackage{tablefootnote}
\usepackage{colortbl}
\usepackage[accsupp]{axessibility}  

\def\wacvPaperID{1778} 
\def\confName{WACV}
\def\confYear{2027}

\title{\vspace{-15pt}
PhotoQuilt: Training-Free Arbitrary-Resolution Photomosaics \\
via Bootstrapped Tiled Denoising
\vspace{-11pt}}

\author{Koorosh Roohi\textsuperscript{1,2,4 *}\quad Javad Rajabi\textsuperscript{1,2,3 *}\quad Andrew Fleet\textsuperscript{2,4,5}\quad Babak Taati\textsuperscript{1,2,4} \\[1mm]
\textsuperscript{1}University of Toronto\quad
\textsuperscript{2}Vector Institute\\
\textsuperscript{3}Samsung Research\quad
\textsuperscript{4}KITE Research Institute\quad
\textsuperscript{5}Queen's University \\
{\tt\small koorosh.roohi@mail.utoronto.ca, rajabi@cs.toronto.edu}\\
\small Project page: \url{https://kooroshrh.github.io/photo-quilt/}
\vspace{-13pt}
}

\begin{document}
\maketitle
\renewcommand{\thefootnote}{\fnsymbol{footnote}}
\footnotetext[1]{Equal contribution.}
\renewcommand{\thefootnote}{\arabic{footnote}}
\renewcommand{\dbltopfraction}{0.9}
\renewcommand{\dblfloatpagefraction}{0.7}

\begin{strip}
  \vspace*{-35pt}\par
  \centering
  \includegraphics[width=0.85\textwidth]{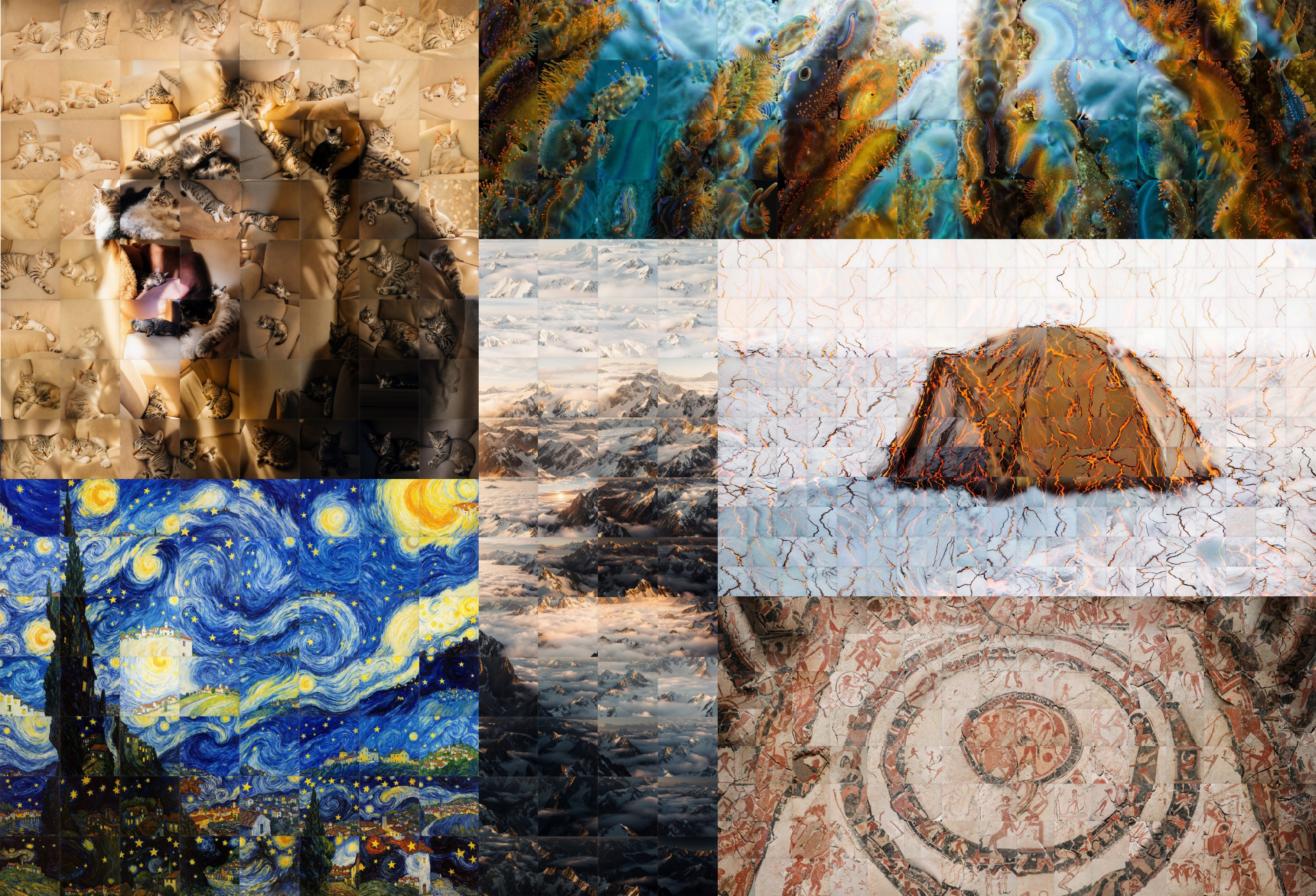}
  \captionof{figure}{High-resolution photomosaics generated across different target images and tile settings. Best viewed zoomed in.}
  \label{fig:teaser}
\end{strip}

\begin{abstract}
Photomosaics are large images whose local regions are seen as independent tiles while their overall arrangement forms a coherent scene. Generating them at high resolution, with every tile convincing in its own right, is computationally expensive, since the canvas must hold many detailed tiles at once. We present \textbf{PhotoQuilt}, a training-free framework that generates photomosaics at arbitrary resolution. Diffusion models struggle to satisfy both scales at once, as direct high-resolution generation is costly and tends toward one smooth image rather than a mosaic, while patch-based tiling keeps local detail but loses global structure. PhotoQuilt resolves this with a bootstrapped tiled denoising procedure. We first produce a global composition at low resolution to fix the layout, then upscale it in latent space and re-inject noise to restore generative capacity. Denoising proceeds within fixed tiles, each re-anchored to its own origin with its local conditioning while sharing the global-level schedule, so that every tile forms its own image rather than a crop of one, while the shared global structure holds them in one layout. Because tile generation is handled separately, PhotoQuilt scales to large canvases without quadratic attention cost. Experiments show that PhotoQuilt provides the best balance between global structure and local realism.
\end{abstract}

\section{Introduction}
\label{sec:intro}

\textit{Mosaics} are among the oldest forms of composite imagery, traditionally constructed by fitting together many small pieces of glass, stone, or ceramic of similar shape and size to form a single larger picture. \textit{Photomosaics} are the digital descendants of this art form, in which the small pieces are themselves independent images, whose collective arrangement reconstructs a second and much larger target image~\cite{silvers1997, ciesielski2007, he2019composing}. What makes a photomosaic visually compelling is that it resolves into two different images depending on the distance from which it is viewed. From up close, the eye is drawn to the detail within each individual tile, every one a self-contained photograph. From afar, this tile-level detail falls away and the target image emerges in its place as a coherent global scene, as shown in~\cref{fig:teaser}.

Photomosaics appear across many fields, from industrial design and advertising to education and digital art~\cite{chung2026photomosaic}. Their meaning comes from how the tiles relate to the larger image they form, so the same technique can express very different ideas, as when photographs from a person's travels come together into a portrait of a place they visited. This expressive power, however, comes with challenges. The same dual-scale behavior that distinguishes a photomosaic from an ordinary image also makes it difficult to produce, since the target image must remain recognizable as a coherent scene from afar while every tile stays a convincing, meaningful image, and satisfying both at the same time is far from trivial. Apart from that, it is also computationally expensive and depends heavily on resolution, since each tile must carry enough detail to read on its own while the canvas holds the many tiles a target image requires, making a faithful mosaic necessarily a high-resolution one and its generation correspondingly costly.

Existing approaches fall into two categories. \emph{Retrieval-based} methods construct a photomosaic by selecting the best-matching tile from a fixed image pool for each region of the target and applying local color corrections to improve the fit~\cite{finkelstein1998}. Because the pool is finite, tiles recur throughout the canvas, and the degree to which corrections can be applied is limited by the need to keep each tile in a good quality, constraining how faithfully tile-level detail and global structure can coexist across viewing scales. \emph{Generative} methods replace retrieval with synthesis, editing, or conditioning each tile on a text prompt or reference image~\cite{podell2024sdxl, zhang2023controlnet, brooks2023instructpix2pix}. Diffusion models make this direction promising, but applying them to photomosaics at high resolution is non-trivial: direct generation is computationally heavy, so most methods generate tiles sequentially~\cite{doyle2026diffusion, chung2026photomosaic}, which is time-consuming and still requires additional coordination procedures to assemble independently generated tiles into a coherent global image.

In this work, we introduce \textit{PhotoQuilt}, a training-free framework that produces photomosaics at arbitrary resolution by decoupling the global composition from the generation of local tiles. Instead of generating tiles in isolation and stitching them together afterward, PhotoQuilt first establishes a coarse global structure and then uses that to guide the generation of every tile. Because every tile is built on this shared foundation, the tiles fit together without a separate step to align them afterward. We begin by producing the target image's layout at low resolution, at very low cost. We then upscale this representation in latent space and re-inject noise. The upscaling enlarges the layout but adds no new detail on its own, so the re-injected noise gives the model the generative capacity to fill in tile-level content, developing each region of the canvas into its own detailed tile. The remaining denoising is carried out within fixed tiles. Each tile is re-anchored to its own origin to develop as an independent image rather than a simple crop by being conditioned separately, while the previously established global structure ensures they combine into a single, coherent scene.

Confining attention within each tile keeps the cost of generation linear in the size of the canvas. This per-tile confinement is what makes high-resolution mosaics practical to produce, yielding outputs in which the target image is clearly recognizable as a whole while every tile remains a meaningful, high-quality image in its own right, a pairing that direct high-resolution generation struggles to achieve. Additionally, since each tile is denoised within its own fixed region of the canvas, the output stays a true mosaic of discrete units instead of blending into a single smooth image. Extensive experiments show that PhotoQuilt provides balanced global coherence and fine-detail fidelity, outperforming existing baselines. Unlike standard high-resolution methods, it achieves this by keeping tiles independent, which makes it a minimal yet effective approach to high-resolution photomosaic generation that remains stable and efficient across a wide range of output resolutions.
\section{Related Works}
\label{sec:related}

\paragraph{Photomosaics and Dual-Scale Composition.} 
A photomosaic is composed of many tile images that, viewed up close, read as distinct photographs, yet collectively reconstruct a target image at a distance. Classical methods~\cite{silvers1997, finkelstein1998} build this by retrieving the best-matching tile from a fixed pool for each block of the target and adjusting its tone~\cite{battiato2006survey}. This approach is inherently constrained by the pool size and limited adjustment options, which lead to repeated tiles and a rigid perceptual scale. A related line of work produces images that read differently at different scales or viewing angles~\cite{geng2024anagrams, geng2024factorized}, though these target a single image with a hidden percept rather than a grid of independently coherent tiles. Chung~\etal~\cite{chung2026photomosaic} introduced a diffusion-based generative photomosaic method, reconstructing a reference image by guiding each tile toward its reference block with a per-step low-frequency loss to balance global structure with local detail, though it remains slow even for modest tile counts. Doyle~\etal~\cite{doyle2026diffusion} take another approach, adapting a text-to-image model and selecting each tile's prompt automatically from the target's colors, which keeps the method training-free. However, this method assembles the mosaic from independently generated tiles and remains costly as the tile count grows. Fast, scalable photomosaic generation on modern generative backbones thus remains an open problem.

\vspace{-7pt}
\paragraph{Diffusion Transformers (DiTs).}
Earlier text-to-image models were built on U-Net backbones, most notably Stable Diffusion~\cite{rombach2022ldm, podell2024sdxl}. Modern text-to-image generation has since shifted to the diffusion transformer (DiT)~\cite{peebles2023dit}, and a rapidly growing family of open DiT foundation models now occupies this space, including SD3~\cite{esser2024sd3}, PixArt-$\alpha$~\cite{chen2024pixart}, Sana~\cite{xie2025sana}, FLUX~\cite{flux2024,flux2_2025}, Qwen-Image~\cite{wu2025qwenimage}, Microsoft Lens~\cite{zhao2026lens}, and the open-weight Ideogram~4~\cite{ideogram2026}. Beyond text-to-image generation, these models increasingly support image-to-image conditioning, in which an input image steers generation through SDEdit-style re-noising~\cite{meng2022sdedit}, image-prompt or image-variation adapters~\cite{ye2023ipadapter,fluxredux2024}, structural control~\cite{zhang2023controlnet}, or unified in-context editing~\cite{labs2025fluxkontext}, with several recent models such as Qwen-Image offering image-to-image generation natively. Our method treats the generator as a black box requiring only text-to-image and image-to-image conditioning, making it compatible in principle with any model in this family, U-Net or DiT, and driving each tile through these interfaces via text or a reference image.


\vspace{-7pt}
\paragraph{Diffusion Model Adaptation.}
Recent work has explored several ways to push pretrained diffusion models beyond their original capabilities. One direction extends the generation to resolutions larger than those seen during training. Patch-based methods denoise overlapping regions and merge them into a single image. Among these, MultiDiffusion~\cite{bartal2023multidiffusion} jointly optimizes all patches, DemoFusion~\cite{du2024demofusion} combines skip residuals with dilated sampling in an upsample, diffuse, and denoise pipeline, and AccDiffusion~\cite{lin2024accdiffusion} adapts text prompts to the content of each patch. These methods differ mainly in how they coordinate neighbouring regions. MultiDiffusion averages overlapping windows at every step, SyncDiffusion~\cite{lee2023syncdiffusion} descends a perceptual loss between their denoised predictions, SpotDiffusion~\cite{frolov2025spotdiffusion} shifts a non-overlapping partition so seams from one step are corrected at the next, and ElasticDiffusion~\cite{hajiali2024elasticdiffusion} splits the trajectory into a global signal at reduced resolution and a local signal on patches. The mechanisms differ, but the purpose is the same, which is to stop neighbouring regions from diverging so the canvas resolves into one continuous image. Later methods improve efficiency, fidelity, and compatibility with DiT backbones~\cite{tragakis2024pixelsmith,qiu2024freescale,kim2025diffusehigh,koh2025scalediff}, while others instead modify inference-time attention or positional encodings to enlarge the model's effective receptive field~\cite{he2024scalecrafter,zhang2024hidiffusion,huang2024fouriscale,rajabi2026sega}. Another line of work conditions pretrained diffusion models on external guidance. ControlNet~\cite{zhang2023controlnet}, T2I-Adapter~\cite{mou2023t2i}, and IP-Adapter~\cite{ye2023ipadapter} train lightweight modules to incorporate structural cues such as edge maps, depth, color palettes, or reference images, while Shum \textit{et al.}~\cite{shum2025color} achieve color palette-guided generation without additional training. The photomosaic problem draws on both threads but fits neither. Like the high-resolution methods, a mosaic must be generated across many tiles to cover a large canvas. Unlike them, it requires neighbouring tiles to diverge, since every tile must read as its own distinct image. Keeping each tile meaningful also requires the flexible conditioning of the second thread, allowing each to follow its own prompt or reference. PhotoQuilt reuses patch-wise generation but lets tiles diverge instead of converge, and conditions each tile on its own.



\section{Method}
\label{sec:method}

We adapt pretrained text-to-image models into zero-shot photomosaic generators. By sharing a coarse latent for the global structure and resolving each tile \emph{independently}, we create images that function at two scales: a cohesive macro image composed of fully self-contained micro images. We guide this process through a dual-control interface, independently defining the global target (real or generated) and the local tile content (prompts or reference images). This method easily scales to high resolutions and massive tile counts (\cref{sec:experiments}).

\subsection{Preliminaries}
\label{subsec:prelim}

\paragraph{Diffusion Models.}
Modern text-to-image models generate images in a compressed latent space by
progressively evolving samples from a pure-noise Gaussian distribution toward a target data distribution through a sequence of intermediate distributions, a process governed by a continuous time
parameter $t \in [0, 1]$. A pretrained encoder $\mathcal{E}$ maps a pixel image
to a latent representation and a decoder $\mathcal{D}$ inverts this mapping, so
that all generation operates in latent space. For a clean latent $z_0$ and
noise $\epsilon \sim \mathcal{N}(0, I)$, the intermediate latent at time $t$ is
\begin{equation}
  z_t = \alpha_t\, z_0 + \sigma_t\, \epsilon,
  \qquad t \in [0,1]
  \label{eq:interp}
\end{equation}
where the schedule coefficients $\alpha_t$ and $\sigma_t$ are chosen so that the
path runs from the clean latent $z_0$ at $t=0$ to pure noise ($\epsilon$) at
$t=1$. Different choices of $\alpha_t$ and $\sigma_t$ recover different
formulations, such as variance-preserving diffusion~\cite{ho2020ddpm, song2020score}
and flow matching~\cite{lipman2023flowmatching, liu2023rectifiedflow, esser2024sd3}.
In our derivation we adopt the linear schedule $\alpha_t = 1-t$ and
$\sigma_t = t$ with $t=0$ the clean latent and $t=1$ pure noise. A network
$v_\theta(z_t, t, c)$ is trained to regress the velocity $\epsilon - z_0$ under a
condition $c$, and sampling runs this process backward in $t$ to recover a clean
latent from noise. We denote by
\begin{equation}
  \Phi\big(z_{\mathrm{init}};\, t_a \!\to\! t_b,\, c\big)
  \label{eq:solver}
\end{equation}
the operation that runs this backward process from $t_a$ to $t_b$, starting at
$z_{\mathrm{init}}$ under condition $c$. 

\vspace{-7pt}
\paragraph{Partial renoising.}
Given a clean latent $z_0$, the standard SDEdit~\cite{meng2022sdedit} operation lets us partially corrupt it with noise and then recover a clean latent again, rather than starting from pure noise. A strength $s \in (0,1)$ controls how far this corruption goes. Using Eq.~\eqref{eq:interp}, we mix $z_0$ with noise up to time $t=s$, and then integrate the backward process from $t=s$ back down to $t=0$ to obtain a new clean latent. SDEdit provides only the shared latent. Because it operates on a single image rather than generating multiple independent outputs, we rely on it strictly to establish global structure, achieving tile autonomy through tiled denoising instead.

\vspace{-7pt}
\paragraph{Tiled denoising.}
A common way to generate large images is to split the latent into spatial tiles, denoise each independently, and recompose them, an approach introduced for patch-wise generation. At high resolution, independently denoised tiles often show repetition. To prevent this, prior methods couple the tiles using techniques like averaging windows~\cite{bartal2023multidiffusion}, perceptual loss~\cite{lee2023syncdiffusion}, shifting partitions~\cite{frolov2025spotdiffusion}, sharing residuals~\cite{du2024demofusion, lin2024accdiffusion}, or injecting global signals~\cite{hajiali2024elasticdiffusion}. We avoid these mechanisms entirely. Because tile divergence is a requirement for our method rather than a flaw, we keep tiles fully independent during denoising and rely solely on a shared initialization for global coherence. This independence requires three modifications to the tiling process: per-tile conditioning, local positional encoding, and a canvas-level timestep shift.

\begin{figure}[pt]
  \centering
  \includegraphics[width=\linewidth]{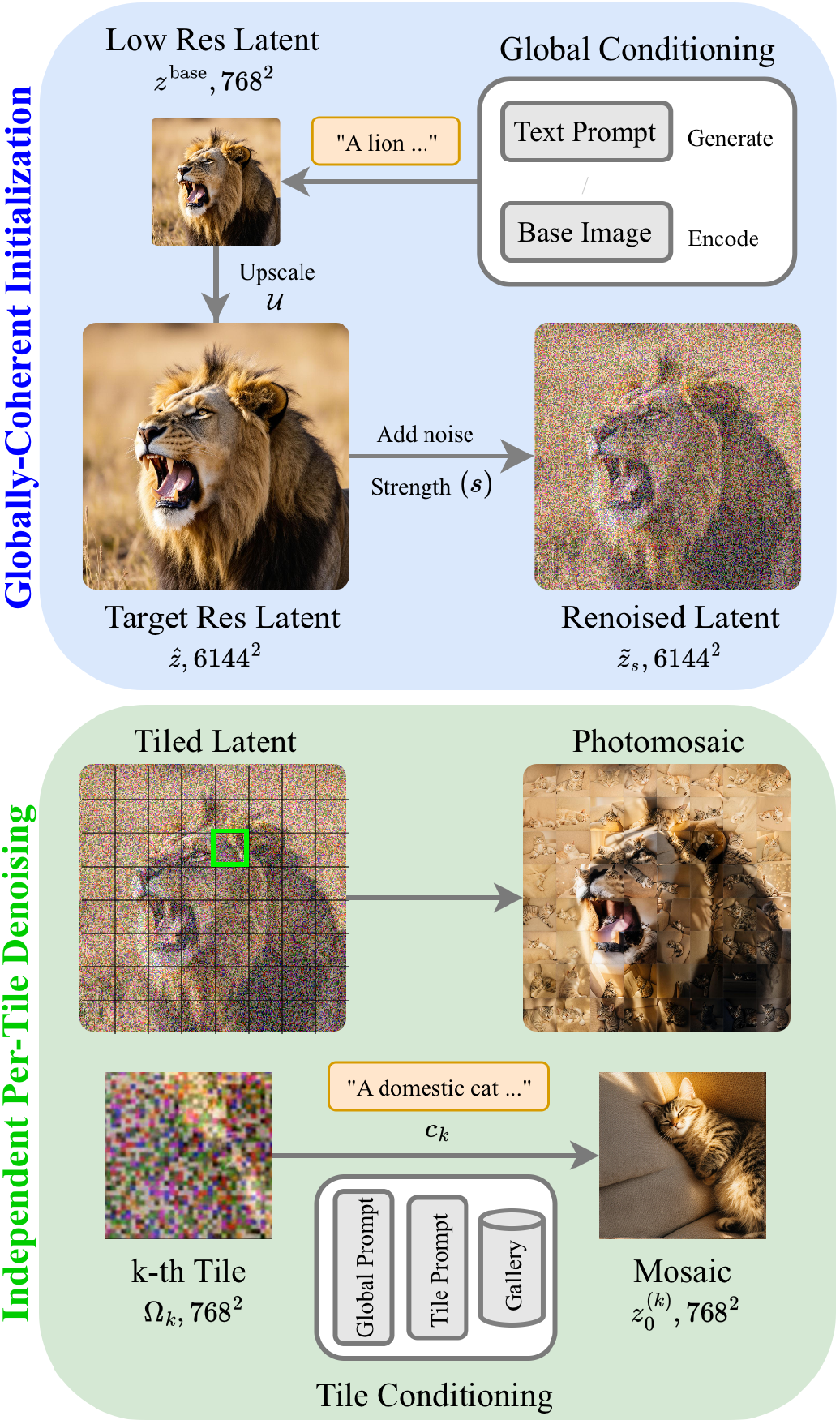}
  \caption{The PhotoQuilt method pipeline. Pixel space representation has been used instead of latent space for simplicity.}
  \label{fig:pipeline}
\end{figure}

\subsection{Problem Formulation and Overview}
\label{subsec:overview}
A photomosaic has a deliberate two-scale structure. At the fine scale, it is a grid of distinct, self-contained images; at the coarse scale, these tiles together reconstruct a single target image. We formalize this two-scale structure with two criteria. Given a global condition $g$, either a base prompt $c_0$ or a reference image $I_0$, a set of $K$ tiles $\{\Omega_k\}_{k=1}^{K}$, and tile conditions $\{c_k\}_{k=1}^{K}$, we seek an image $I^{\mathrm{mosaic}}$ at the target resolution such that:

\begin{enumerate}
  \item[(i)] \textbf{Global Reconstruction.} The low-frequency content of
  $I^{\mathrm{mosaic}}$ reconstructs the target structure specified by $g$.

  \item[(ii)] \textbf{Tile Autonomy.} Each tile ($\Omega_k$) is a complete,
  self-contained image consistent with $c_k$.
\end{enumerate}

Criterion (ii) separates the photomosaic objective from standard patch-wise high-resolution generations. Consequently, Sec. F of supplemental directly compares our overlap padding against MultiDiffusion-style window averaging at a matched overlap width. We achieve these two criteria with two separate mechanisms. A single coarse latent, shared across all tiles and renoised once, fixes the low-frequency layout they have in common, giving us global reconstruction. Each tile is denoised as its own independent trajectory from that shared latent, giving us tile autonomy.

\subsection{Bootstrapped Tiled Denoising}
\paragraph{Globally-Coherent Initialization.}
\label{subsec:init}

We first obtain a base latent $z^{\mathrm{base}}$ at a low resolution, either by generating it from the base prompt ($c_0$) or by encoding a provided image ($I_0$),
\begin{equation}
  z^{\mathrm{base}} =
  \begin{cases}
    \Phi\big(\epsilon^{\mathrm{low}};\, 1\!\to\! 0,\, c_0\big),
      & \text{(generated base)}\\[2pt]
    \mathcal{E}(I_0),
      & \text{(image base)}
  \end{cases}
  \label{eq:base}
\end{equation}
with $\epsilon^{\mathrm{low}}\sim\mathcal{N}(0,I)$ and $\mathcal{E}$ as the encoder. We upsample that to the target latent grid using a fixed upsampler $\mathcal{U}$, then renoise it once with strength $s$,
\begin{equation}
  \hat{z} = \mathcal{U}\big(z^{\mathrm{base}}\big),
  \qquad
  \tilde{z}_s = (1-s)\,\hat{z} + s\,\epsilon,
  \quad \epsilon\sim\mathcal{N}(0,I)
  \label{eq:renoise}
\end{equation}
Because $s<1$, the renoised latent $\tilde{z}_s$ retains the coarse content of $\hat{z}$ while leaving its high-frequency detail to be regenerated. As $\tilde{z}_s$ is shared by every tile, it fixes a common low-frequency layout across the image. The strength $s$ thus acts as a single global/local control: smaller $s$ enforces the global structure more strictly, while larger $s$ grants each tile more freedom to diverge. 

\vspace{-7pt}
\paragraph{Independent Per-Tile Denoising.}
\label{subsec:tiles}

We split the target latent into $K$ equal, non-overlapping tiles
$\{\Omega_k\}$ and denoise all tiles together, each as a separate trajectory. Each tile starts from its own region of the shared renoised latent and follows its own condition $c_k$,
\begin{equation}
  z_0^{(k)} =
  \Phi\Big(\, \tilde{z}_s\big|_{\Omega_k};\; s\!\to\! 0,\; c_k \Big),
  \qquad k = 1,\dots,K
  \label{eq:tile}
\end{equation}
where $\tilde{z}_s\big|_{\Omega_k}$ restricts the shared latent to tile $\Omega_k$; denoising runs at native resolution per tile. Each tile carries its own copy of the condition tokens, positioned at the tile origin unlike patch-wise high resolution approaches, and attention is confined within the tile, so Eq.~\eqref{eq:tile} is exactly separable across $K$. Tiles can be evaluated in any order, batched, or distributed across devices with no communication between them, and the cost of the canvas is linear in $K$ rather than quadratic in its area. Independence is not obtained by simply running $K$ separate generations: the timestep shift of the denoising schedule is computed once from the full-canvas sequence length and shared by all tiles, since computing it per tile places each tile on a different trajectory. The supplementary materials explains the distributed mechanism in Sec. E of supplemental. However, initial latent of each tile can be an encoded image just like using a image base for the global appearance, or it can be conditioned using an image instead of text prompt. These different modes are explained in Sec. B of supplementary materials.

\vspace{-7pt}
\paragraph{Final Photomosaic.}
We now decode the denoised global latent, consisting of independently denoised tiles,
\begin{equation}
  I^{\mathrm{mosaic}} =
  \mathcal{D}\big(\, \{z_0^{(k)}\}_{k=1}^{K} \,\big)
  \label{eq:decode}
\end{equation}
Tile independence in Eq.~\eqref{eq:tile} is what makes the output a mosaic rather than an upsampled image. By default, tiles do not overlap, so the seams between tiles are obvious. When a smoother, continuous appearance is preferred, the tiles can be given a small overlap and blended where they meet, which removes visible seams. This is optional and not used in our main results; this is further explained in Sec. F of supplemental. See~\cref{fig:pipeline} for the PhotoQuilt pipeline.

\begin{algorithm}[t]
\caption{PhotoQuilt: bootstrapped tiled denoising}
\label{alg:main}
\begin{algorithmic}[1]
\Require global condition $g$; tile conditions
         $\{c_k\}_{k=1}^{K}$; strength $s$; partition
         $\{\Omega_k\}_{k=1}^{K}$; fixed $\mathcal{E},\mathcal{D},\mathcal{U}$
\Ensure  photomosaic image $I^{\mathrm{mosaic}}$
\State $z^{\mathrm{base}} \gets
       (\Phi(\epsilon^{\mathrm{low}};\,1\!\to\!0,\,c_0)$ \textbf{ or } $\mathcal{E}(I_0))$
       \Comment{Eq.~\eqref{eq:base}}
\State $\hat{z} \gets \mathcal{U}(z^{\mathrm{base}})$
       \Comment{upsample to target grid}
\State $\tilde{z}_s \gets (1-s)\,\hat{z} + s\,\epsilon, \;\;
       \epsilon \sim \mathcal{N}(0,I)$
       \Comment{Eq.~\eqref{eq:renoise}}
\ForAll{$k \in \{1,\dots,K\}$ \textbf{(parallel)}}
  \State $z_0^{(k)} \gets \Phi\!\left(\tilde{z}_s\big|_{\Omega_k};\;
         s\!\to\!0,\; c_k\right)$
         \Comment{Eq.~\eqref{eq:tile}}
\EndFor
\State \Return $I^{\mathrm{mosaic}} \gets
       \mathcal{D}\!\left(\{z_0^{(k)}\}_{k=1}^{K}\right)$
       \Comment{Eq.~\eqref{eq:decode}}
\end{algorithmic}
\end{algorithm}
\vspace{-5pt}
\section{Experiments}
\label{sec:experiments}
\subsection{Experimental Setup}
\paragraph{Implementation Details.} We evaluate the model-agnostic PhotoQuilt across three backbone configurations: Stable Diffusion 2.1 (SD2.1)~\cite{rombach2022ldm}, FLUX.1~\cite{flux2024}, and FLUX.2~\cite{flux2_2025}. The SD2.1 backbone is included to enable a fair comparison with competing methods that are constrained to this backbone. All methods operate on mosaics of resolution $6144 \times 6144$ with tile size $768 \times 768$ (8 tiles per axis), matching the native generation resolution of SD2.1. To ensure a consistent global structure across all evaluated methods, the base image for every method is generated at $768 \times 768$ using FLUX.1 from the same prompt, giving all methods an identical starting point. Bicubic interpolation has been used as the upscaling function. To construct our evaluation set, we selected 50 prompts from the Aesthetic-4K~\cite{zhang2025diffusion} dataset that highlight distinct photomosaic characteristics, pairing each with 15 random seeds to generate 750 unique samples.


\vspace{-7pt}
\paragraph{Metrics.} 
We evaluate global structure and local tile quality using two complementary sets of metrics. Global structure is measured by comparing a downsampled mosaic to the $64 \times 64$ base image (see Sec. C of supplemental for more resolutions) via PSNR, SSIM~\cite{wang2004image}, LPIPS~\cite{zhang2018unreasonable}, HPSv2~\cite{wu2024hps}, Image Reward (IR)~\cite{xu2023imagereward}, and FID~\cite{fid}. Global FID lacks standard deviation in the tables due to its nature. Local tile quality is evaluated on individual tiles for prompt alignment and fidelity using CLIP~\cite{radford2021clip}, BLIP~\cite{li2022blip}, CLIP-IQA~\cite{wang2023clipiqa}, HPSv2~\cite{wu2024hps}, IR~\cite{xu2023imagereward}, and FID~\cite{fid}. Finally, we introduce the Dual-scale Reward (DSR) which is the harmonic mean of the global and local IR scores as a unified metric for overall performance and evaluating both global and tile scales. This metric is further explained in Sec. I of supplementary materials.

\begin{figure*}[pt!]
  \centering
  \includegraphics[width=0.95\textwidth]{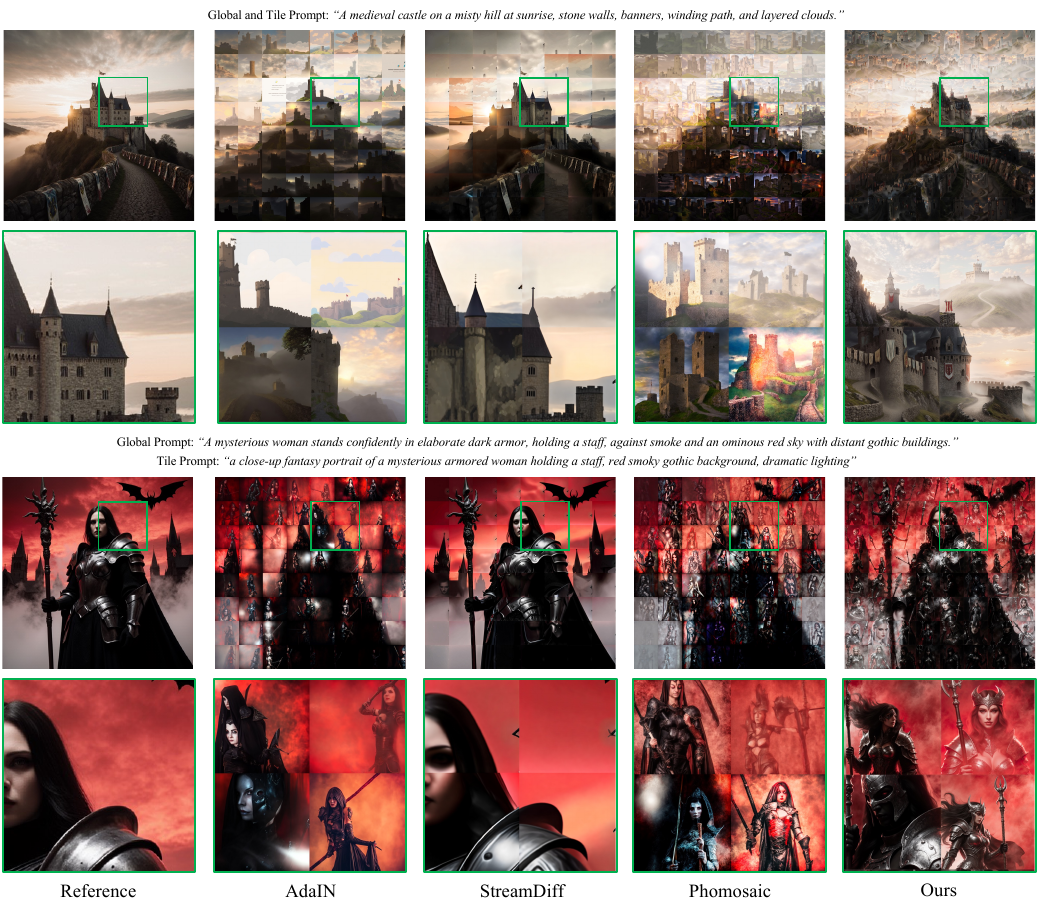}
  \caption{Qualitative comparison of photomosaic generation. Compared to baselines, PhotoQuilt better preserves the global target structure while generating realistic, self-contained tiles.}
  \label{fig:qual-comp}
\end{figure*}

\subsection{Baseline Methods}
We compare PhotoQuilt against six baselines. \textit{Match \& Tone}~\cite{silvers1997,finkelstein1998} is the classical retrieval pipeline: each block is matched to a fixed image pool and its tone is adjusted to the target, serving as our non-generative lower bound. \textit{AdaIN}~\cite{huang2017adain} transfers the color and style statistics of each target block onto an independently generated tile, lacking explicit structural conditioning. \textit{Color T2I-Adapter}~\cite{mou2023t2i} injects a block-level color map via an adapter network as a spatial conditioning signal during tile denoising. \textit{NoiseBlend}~\cite{lee2024diffusion} combines each tile's initial noise with a crop of the target latent at a fixed ratio, approximating our shared initialization but without explicit renoising. \textit{StreamDiff}~\cite{kodaira2023streamdiffusion} processes all tiles as a high-throughput batch under the same prompt with no global coordination. \textit{Phomosaic}~\cite{chung2026photomosaic} guides SD2.1 tile generation with a per-step low-frequency structural loss and AdaIN-style color alignment. More details regarding the implementation of these methods are provided in Sec. A of the supplementary materials.


\begin{table*}[t]
\centering
\caption{Same-backbone comparison and backbone scaling, mean (std) over all samples. (a) All methods on SD2.1, identical settings. (b) PhotoQuilt across backbones. Global structure at $64\times64$. Bold/underline = best/second-best \emph{within each subgroup}. Tiled high-resolution methods (MultiDiffusion, SpotDiffusion, ScaleDiff, ElasticDiffusion) are bounded by StreamDiff on criterion (ii); see Sec. 2 and Sec. F.}
\label{tab:main}
\resizebox{\textwidth}{!}{
\scriptsize
\begin{tabular}{l cccccc cccccc c}
\toprule
\multirow{2}{*}{\textbf{Method}} & \multicolumn{6}{c}{\textbf{Global Structure ($64\times64$)}} & \multicolumn{6}{c}{\textbf{Local Tiles}} & \multirow{2}{*}{\textbf{DSR} $\uparrow$} \\
\cmidrule(lr){2-7} \cmidrule(lr){8-13}
& PSNR $\uparrow$ & SSIM $\uparrow$ & LPIPS $\downarrow$ & HPS $\uparrow$ & IR $\uparrow$ & FID $\downarrow$ & CLIP $\uparrow$ & BLIP $\uparrow$ & IQA $\uparrow$ & HPS $\uparrow$ & IR $\uparrow$ & FID $\downarrow$ & \\
\midrule
\multicolumn{14}{l}{\textbf{(a) Same-backbone comparison (SD2.1)}} \\[1pt]
Match \& Tone & 15.68\sd{2.10} & 0.27\sd{0.10} & 0.370\sd{0.095} & 0.117\sd{0.022} & -2.27\sd{0.03} & 202.97 & 0.330\sd{0.031} & 0.78\sd{0.23} & 0.79\sd{0.12} & 0.204\sd{0.032} & -0.58\sd{0.73} & 338.59\sd{36.0} & 0.148 \\
AdaIN & 19.76\sd{1.91} & 0.58\sd{0.08} & 0.215\sd{0.077} & 0.136\sd{0.029} & -2.10\sd{0.42} & 199.31 & \underline{0.345}\sd{0.031} & \underline{0.88}\sd{0.16} & 0.63\sd{0.17} & 0.200\sd{0.032} & \textbf{-0.30}\sd{0.65} & \textbf{287.21}\sd{24.9} & \underline{0.174} \\
Color T2I-Adapter & 10.97\sd{1.20} & 0.14\sd{0.04} & 0.572\sd{0.097} & 0.114\sd{0.022} & -2.27\sd{0.03} & 327.78 & 0.334\sd{0.027} & 0.77\sd{0.20} & \underline{0.85}\sd{0.10} & \underline{0.209}\sd{0.033} & -0.59\sd{0.68} & 295.39\sd{31.5} & 0.148 \\
NoiseBlend & 14.67\sd{1.82} & 0.31\sd{0.10} & 0.298\sd{0.088} & 0.216\sd{0.034} & -0.38\sd{1.06} & \textbf{125.35} & 0.151\sd{0.043} & 0.06\sd{0.11} & 0.23\sd{0.06} & 0.071\sd{0.019} & -2.24\sd{0.06} & 363.19\sd{33.3} & 0.155 \\
StreamDiff & \underline{26.37}\sd{3.57} & \underline{0.89}\sd{0.06} & \textbf{0.032}\sd{0.029} & \underline{0.243}\sd{0.035} & \textbf{0.20}\sd{1.09} & \underline{135.92} & 0.201\sd{0.032} & 0.17\sd{0.13} & 0.36\sd{0.09} & 0.100\sd{0.016} & -2.02\sd{0.21} & 310.99\sd{31.8} & \underline{0.193} \\
Phomosaic & 20.77\sd{1.45} & 0.57\sd{0.05} & 0.237\sd{0.067} & 0.155\sd{0.023} & -2.14\sd{0.30} & 222.12 & \textbf{0.346}\sd{0.023} & 0.87\sd{0.14} & 0.79\sd{0.09} & 0.208\sd{0.030} & \underline{-0.33}\sd{0.51} & 297.53\sd{32.6} & 0.168 \\
\textbf{PhotoQuilt (SD2.1)} & \textbf{27.46}\sd{1.24} & \textbf{0.91}\sd{0.05} & \underline{0.082}\sd{0.036} & \textbf{0.257}\sd{0.029} & \underline{0.05}\sd{0.76} & 146.55 & \textbf{0.346}\sd{0.028} & \textbf{0.92}\sd{0.19} & \textbf{0.86}\sd{0.06} & \textbf{0.215}\sd{0.027} & -0.74\sd{0.54} & \underline{294.25}\sd{34.3} & \textbf{0.397} \\
\midrule
\multicolumn{14}{l}{\textbf{(b) Backbone scaling (PhotoQuilt)}} \\[1pt]
\rowcolor{gray!10} PhotoQuilt (SD2.1) & 27.46\sd{1.24} & \textbf{0.91}\sd{0.05} & \underline{0.082}\sd{0.036} & \textbf{0.257}\sd{0.029} & 0.05\sd{0.76} & \textbf{146.55} & 0.346\sd{0.028} & \underline{0.92}\sd{0.19} & \textbf{0.86}\sd{0.06} & 0.215\sd{0.027} & -0.74\sd{0.54} & \underline{294.25}\sd{34.3} & 0.397 \\
PhotoQuilt (FLUX.1) & \textbf{29.60}\sd{1.74} & \textbf{0.91}\sd{0.04} & \textbf{0.035}\sd{0.034} & \underline{0.238}\sd{0.034} & \textbf{0.17}\sd{0.98} & \underline{187.69} & \textbf{0.354}\sd{0.028} & \textbf{0.95}\sd{0.06} & 0.81\sd{0.12} & \textbf{0.231}\sd{0.031} & \textbf{0.25}\sd{0.64} & 296.75\sd{29.2} & \textbf{0.552} \\
PhotoQuilt (FLUX.2) & \underline{28.71}\sd{1.31} & \underline{0.64}\sd{0.06} & 0.188\sd{0.062} & \underline{0.238}\sd{0.034} & \underline{0.08}\sd{0.80} & 200.00 & \underline{0.352}\sd{0.028} & 0.87\sd{0.13} & \underline{0.82}\sd{0.07} & \underline{0.228}\sd{0.030} & \underline{0.11}\sd{0.45} & \textbf{285.12}\sd{27.9} & \underline{0.524} \\
\bottomrule
\end{tabular}
}
\end{table*}

\subsection{Evaluation}
\paragraph{Quantitative results.}
Results are shown in Table~\ref{tab:main}. A key observation holds across all baselines: global fidelity and local tile quality trade off against each other, and no prior method achieves both. StreamDiff achieves the highest PSNR and SSIM among baselines by processing all tiles under the same image-to-image stream, but this comes at the cost of tile diversity as tiles fail to diverge and instead produce near-identical content, which collapses local CLIP, BLIP, and preference scores. The color-conditioning methods (Color T2I-Adapter, AdaIN, Phomosaic) maintain reasonable tile-level scores but sacrifice global structure fidelity, as they transfer only low-level color statistics to each tile rather than adapting tile content. Match \& Tone achieves moderate local scores, but fails on global metrics.

PhotoQuilt breaks this trade-off across all tested backbones. On SD2.1, it outperforms Phomosaic in global structure while matching or exceeding its tile quality except on IR. This is also supported by the DSR metric which evaluates the photomosaic effect that PhotoQuilt substantially outperforms baselines. Critically, this improvement stems from content-level coordination rather than simple color transfer; the shared renoised base provides a spatially-aware initialization, forcing tiles to adapt their distinct content to fit the global target (Fig.~\ref{fig:qual-comp}). Furthermore, our training-free procedure is backbone-agnostic and with a variety of backbones: PhotoQuilt (FLUX.1) achieves the highest CLIP and IR scores alongside strong global metrics, while FLUX.2 posts the strongest tile FID scores.

\paragraph{Qualitative results.}
Fig.~\ref{fig:qual-comp} compares methods across two prompts with varying global structures and conditioning modes. While AdaIN and Phomosaic preserve approximate color distributions, their individual tiles ignore global spatial arrangements; the zoomed insets show plausible but completely disconnected images. Conversely, StreamDiff faithfully reproduces the global image but sacrifices tile autonomy, reducing the mosaic to a single blurry image rather than distinct photographs. Our method successfully bridges this gap, generating self-contained tiles that coherently construct the global composition across all two scenes.


\begin{figure*}[pt!]
  \centering
  \includegraphics[width=0.95\textwidth]{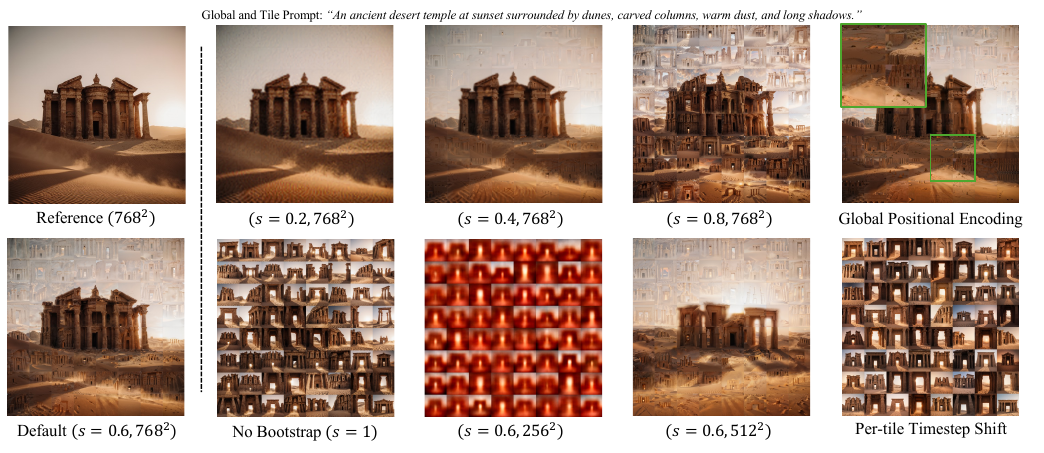}
  \caption{The default configuration ($s=0.6$, $768\times768$) provides the best balance across global layout fidelity and local tile realism compared to the ablated variants.}
  \label{fig:qual-comp-ab}
\end{figure*}

\vspace{-5pt}
\begin{table*}[t]
\centering
\caption{Ablation study on the FLUX.1 backbone, and on a 12-prompt subset: renoising strength ($s$), base image (bootstrap), and the three essential mechanisms that make tiles independent and photomosaic effect possible.}
\label{tab:ablation_study}
\newcommand{\good}[1]{{\scriptsize\color{green!60!black} (#1)}}
\newcommand{\bad}[1]{{\scriptsize\color{red!80!black} (#1)}}
\newcommand{\neut}[1]{{\scriptsize\color{gray} (#1)}}
\resizebox{\textwidth}{!}{
\begin{tabular}{l ccccc ccccc c}
\toprule
\multirow{2}{*}{\textbf{Configuration}} & \multicolumn{5}{c}{\textbf{Global Structure ($64\times64$)}} & \multicolumn{5}{c}{\textbf{Local Tile Quality}} & \multirow{2}{*}{\textbf{DSR} $\uparrow$} \\
\cmidrule(lr){2-6} \cmidrule(lr){7-11}
& PSNR $\uparrow$ & SSIM $\uparrow$ & LPIPS $\downarrow$ & HPS $\uparrow$ & IR $\uparrow$ & CLIP $\uparrow$ & BLIP $\uparrow$ & IQA $\uparrow$ & HPS $\uparrow$ & IR $\uparrow$ & \\
\midrule
\textbf{Default} ($s=0.6$, $768^2$) & \textbf{30.15} & \textbf{0.89} & \textbf{0.04} & \textbf{0.23} & \textbf{0.09} & \textbf{0.36} & \textbf{0.93} & \textbf{0.79} & \textbf{0.23} & \textbf{0.21} & \textbf{0.537} \\
\midrule
\multicolumn{12}{l}{\textbf{Renoising Strength} (Bootstrap = $768\times768$)} \\
\midrule
$s = 0.2$ & 42.55 \good{+12.40} & 0.99 \good{+0.10} & 0.00 \good{-0.04} & 0.25 \good{+0.02} & 0.59 \good{+0.50} & 0.16 \bad{-0.20} & 0.02 \bad{-0.91} & 0.38 \bad{-0.41} & 0.08 \bad{-0.15} & -2.27 \bad{-2.48} & 0.163 \bad{-0.374} \\
$s = 0.4$ & 36.55 \good{+6.40} & 0.97 \good{+0.08} & 0.00 \good{-0.04} & 0.24 \good{+0.01} & 0.54 \good{+0.45} & 0.31 \bad{-0.05} & 0.68 \bad{-0.25} & 0.50 \bad{-0.29} & 0.15 \bad{-0.08} & -1.24 \bad{-1.45} & 0.331 \bad{-0.206} \\
$s = 0.8$ & 22.71 \bad{-7.44} & 0.62 \bad{-0.27} & 0.24 \bad{+0.20} & 0.18 \bad{-0.05} & -1.49 \bad{-1.58} & 0.37 \good{+0.01} & 0.97 \good{+0.04} & 0.85 \good{+0.06} & 0.26 \good{+0.03} & 0.76 \good{+0.55} & 0.290 \bad{-0.247} \\
\midrule
\multicolumn{12}{l}{\textbf{Bootstrap Size} ($s = 0.6$)} \\
\midrule
Bootstrap = $256\times256$ & 9.88 \bad{-20.27} & 0.03 \bad{-0.86} & 0.52 \bad{+0.48} & 0.08 \bad{-0.15} & -2.28 \bad{-2.37} & 0.16 \bad{-0.20} & 0.02 \bad{-0.91} & 0.29 \bad{-0.50} & 0.07 \bad{-0.16} & -2.27 \bad{-2.48} & 0.093 \bad{-0.444} \\
Bootstrap = $512\times512$ & 29.03 \bad{-1.12} & 0.87 \bad{-0.02} & 0.04 \neut{$\pm0.00$} & 0.23 \neut{$\pm0.00$} & -0.05 \bad{-0.14} & 0.36 \neut{$\pm0.00$} & 0.92 \bad{-0.01} & 0.78 \bad{-0.01} & 0.23 \neut{$\pm0.00$} & 0.18 \bad{-0.03} & 0.515 \bad{-0.022} \\
\midrule
\multicolumn{12}{l}{\textbf{Other Ablations}} \\
\midrule
Without Bootstrap & 13.10 \bad{-17.05} & 0.11 \bad{-0.78} & 0.54 \bad{+0.50} & 0.12 \bad{-0.11} & -2.28 \bad{-2.37} & 0.37 \good{+0.01} & 1.00 \good{+0.07} & 0.88 \good{+0.09} & 0.29 \good{+0.06} & 1.15 \good{+0.94} & 0.165 \bad{-0.372} \\
Global Positional Encoding & 31.24 \good{+1.09} & 0.88 \bad{-0.01} & 0.06 \bad{+0.02} & 0.24 \good{+0.01} & 0.09 \neut{$\pm0.00$} & 0.17 \bad{-0.19} & 0.70 \bad{-0.23} & 0.48 \bad{-0.31} & 0.14 \bad{-0.09} & -0.05 \bad{-0.26} & 0.504 \bad{-0.033} \\
Per-tile Timestep Shift & 14.98 \bad{-15.17} & 0.13 \bad{-0.76} & 0.49 \bad{+0.45} & 0.11 \bad{-0.12} & -1.98 \bad{-2.07} & 0.36 \neut{$\pm0.00$} & 1.00 \good{+0.07} & 0.89 \good{+0.10} & 0.28 \good{+0.05} & 1.09 \good{+0.88} & 0.209 \bad{-0.328} \\
\bottomrule
\end{tabular}
}
\end{table*}

\subsection{Ablation Study}
We ablate the design choices and mechanisms of PhotoQuilt on the FLUX.1 backbone on a subset of 12 prompts from the main dataset, reported in Table~\ref{tab:ablation_study}. All configurations use the $6144\times6144$ target resolution with $64$ tiles.

\vspace{-9pt}
\paragraph{Global guidance (bootstrap).}
Removing the bootstrapped initialization entirely and replacing the shared renoised latent with independent per-tile noise, maximize local tile quality: BLIP reaches $1.00$ and Image Reward improves by $+0.94$, as each tile is free to denoise from scratch toward its condition without any structural constraint. However, global structure collapses, with PSNR dropping to $13.10$ and SSIM to $0.11$, confirming that the shared renoised base is the sole source of layout coherence.

\vspace{-8pt}
\paragraph{Renoising strength.}
The strength $s \in (0, 1)$ controls what fraction of the full denoising trajectory is reserved for tile-level generation: small $s$ injects little noise and preserves the coarse latent almost intact, while large $s$ approaches full renoising and grants each tile near-complete generative freedom. The ablation sweeps $s \in \{0.2, 0.4, 0.6, 0.8\}$ and reveals a clean monotonic trade-off, as shown in~\cref{fig:qual-comp-ab}. At $s=0.2$, global fidelity peaks (PSNR $42.55$, SSIM $0.99$) but tiles are forced to complete an almost clean latent, leaving little room to diverge; local scores degrade sharply, with Image Reward falling to $-2.27$. Increasing $s$ relaxes this constraint: at $s=0.8$, tile-level Image Reward rises to $0.76$ but global SSIM falls to $0.62$. The default $s=0.6$ sits at the crossover where both criteria are simultaneously well-satisfied, achieving the best balance between layout fidelity and tile independence.

\vspace{-10pt}
\paragraph{Bootstrap resolution.}
We vary the resolution of the base image generated in the first stage before upsampling, testing $256\!\times\!256$, $512\!\times\!512$, and the default $768\!\times\!768$. At $256\times256$ the global structure collapses entirely (PSNR $9.88$, SSIM $0.03$), indicating that a small bootstrap retains too little spatial detail for the tiles to recover the global layout. At $512\!\times\!512$ the degradation is mild (PSNR $-1.12$, SSIM $-0.02$). The full $768\!\times\!768$ bootstrap consistently provides the richest structural guidance and is the default for all reported results. Together, these results show that bootstrap quality propagates directly into global fidelity: a higher-resolution base encodes more spatial detail into $\tilde{z}_s$, giving each tile a more informative initialization of its region of the global image.

\vspace{-8pt}
\paragraph{Tile independence mechanisms.}
Two further ablations demonstrate what separates our approach from naive tiled denoising. First, removing local positional encoding preserves global structure but ruins tile quality, turning tiles into mere crops of a single image rather than self-contained photographs. Second, computing the timestep shift per-tile places each on a different trajectory; this completely breaks global coherence (PSNR $14.98$, SSIM $0.13$) while artificially inflating individual tile scores. Therefore, both mechanisms are essential, each satisfying a distinct criterion.
\vspace{-3pt}
\section{Conclusion}
We presented PhotoQuilt, a training-free framework for generating photomosaics at arbitrary resolution. Instead of generating tiles in isolation and combining them afterward, it decouples global composition from local generation through bootstrapped tiled denoising, fixing a coarse layout at low resolution, upscaling it in latent space, and renoising it once so that a single shared latent enforces global structure while each tile denoises as its own trajectory. Since tiles are generated separately, the method scales to large canvases without quadratic attention cost. Requiring no training or architectural change, PhotoQuilt applies to both U-Net and DiT backbones, providing the best balance between global structure and local realism. Remaining challenges are in image-gallery conditioning mode: the quality is tied to the backbone, because a denoised tile can drift away from its origin, and there are also cases of globally over- or under-constrained canvases, such as landscapes, portraits, or text where photomosaic effect is troubled.

{
    \small
    \bibliographystyle{ieeenat_fullname}
    \bibliography{main}

@String(CVPR  = {CVPR})

@String(ICCV  = {ICCV})

@String(ECCV  = {ECCV})

@String(ICLR  = {ICLR})

@String(AAAI  = {AAAI})

@inproceedings{liu2023rectifiedflow,
  title     = {Flow Straight and Fast: Learning to Generate and Transfer
               Data with Rectified Flow},
  author    = {Liu, Xingchao and Gong, Chengyue and Liu, Qiang},
  booktitle = {The Eleventh International Conference on Learning
               Representations (ICLR)},
  year      = {2023},
  url       = {https://openreview.net/forum?id=XVjTT1nw5z}
}

@inproceedings{lipman2023flowmatching,
  title     = {Flow Matching for Generative Modeling},
  author    = {Lipman, Yaron and Chen, Ricky T. Q. and Ben-Hamu, Heli
               and Nickel, Maximilian and Le, Matt},
  booktitle = {The Eleventh International Conference on Learning
               Representations (ICLR)},
  year      = {2023},
  url       = {https://openreview.net/forum?id=PqvMRDCJT9t}
}

@inproceedings{esser2024sd3,
  title     = {Scaling Rectified Flow Transformers for High-Resolution
               Image Synthesis},
  author    = {Esser, Patrick and Kulal, Sumith and Blattmann, Andreas
               and Entezari, Rahim and M{\"u}ller, Jonas and Saini, Harry
               and Levi, Yam and Lorenz, Dominik and Sauer, Axel and
               Boesel, Frederic and Podell, Dustin and Dockhorn, Tim and
               English, Zion and Rombach, Robin},
  booktitle = {Proceedings of the 41st International Conference on
               Machine Learning (ICML)},
  series    = {Proceedings of Machine Learning Research},
  volume    = {235},
  pages     = {12606--12633},
  year      = {2024},
  url       = {https://proceedings.mlr.press/v235/esser24a.html}
}

@misc{flux2024,
  author       = {{Black Forest Labs}},
  title        = {{FLUX}: Text-to-Image Generation Model},
  year         = {2024},
  howpublished = {\url{https://github.com/black-forest-labs/flux}},
  note         = {Released August 2024}
}

@misc{flux2_2025,
  author       = {{Black Forest Labs}},
  title        = {{FLUX.2}: Frontier Visual Intelligence},
  year         = {2025},
  howpublished = {\url{https://github.com/black-forest-labs/flux2}},
  note         = {Released November 2025}
}

@misc{fluxredux2024,
  author       = {{Black Forest Labs}},
  title        = {{FLUX.1} Tools: {Redux}, Fill, Depth, Canny},
  year         = {2024},
  howpublished = {\url{https://bfl.ai/flux-1-tools/}},
  note         = {Released November 2024}
}

@inproceedings{meng2022sdedit,
  title     = {{SDEdit}: Guided Image Synthesis and Editing with
               Stochastic Differential Equations},
  author    = {Meng, Chenlin and He, Yutong and Song, Yang and Song,
               Jiaming and Wu, Jiajun and Zhu, Jun-Yan and Ermon,
               Stefano},
  booktitle = {International Conference on Learning Representations
               (ICLR)},
  year      = {2022},
  url       = {https://openreview.net/forum?id=aBsCjcPu_tE}
}

@inproceedings{bartal2023multidiffusion,
  title     = {{MultiDiffusion}: Fusing Diffusion Paths for Controlled
               Image Generation},
  author    = {Bar-Tal, Omer and Yariv, Lior and Lipman, Yaron and
               Dekel, Tali},
  booktitle = {Proceedings of the 40th International Conference on
               Machine Learning (ICML)},
  series    = {Proceedings of Machine Learning Research},
  volume    = {202},
  pages     = {1737--1752},
  year      = {2023},
  url       = {https://proceedings.mlr.press/v202/bar-tal23a.html}
}

@inproceedings{du2024demofusion,
  title     = {{DemoFusion}: Democratising High-Resolution Image
               Generation With No {{\$\$\$}}},
  author    = {Du, Ruoyi and Chang, Dongliang and Hospedales, Timothy
               and Song, Yi-Zhe and Ma, Zhanyu},
  booktitle = {Proceedings of the IEEE/CVF Conference on Computer
               Vision and Pattern Recognition (CVPR)},
  pages     = {6159--6168},
  year      = {2024}
}

@inproceedings{lin2024accdiffusion,
  title     = {{AccDiffusion}: An Accurate Method for Higher-Resolution
               Image Generation},
  author    = {Lin, Zhihang and Lin, Mingbao and Zhao, Meng and Ji,
               Rongrong},
  booktitle = {European Conference on Computer Vision (ECCV)},
  year      = {2024},
  url       = {https://arxiv.org/abs/2407.10738}
}

@book{silvers1997,
  author    = {Silvers, Robert and Hawley, Michael},
  title     = {Photomosaics},
  publisher = {Henry Holt and Co.},
  year      = {1997}
}

@inproceedings{finkelstein1998,
  author    = {Finkelstein, Adam and Range, Marisa},
  title     = {Image Mosaics},
  booktitle = {Electronic Publishing, Artistic Imaging, and Digital
               Typography (RIDT)},
  series    = {Lecture Notes in Computer Science},
  volume    = {1375},
  pages     = {11--22},
  publisher = {Springer},
  year      = {1998}
}

@inproceedings{battiato2006survey,
  author    = {Battiato, Sebastiano and Di Blasi, Gianpiero and
               Farinella, Giovanni Maria and Gallo, Giovanni},
  title     = {A Survey of Digital Mosaic Techniques},
  booktitle = {Eurographics Italian Chapter Conference},
  year      = {2006}
}

@article{chung2026photomosaic,
  title   = {Generative Phomosaic with Structure-Aligned and
             Personalized Diffusion},
  author  = {Chung, Jaeyoung and Son, Hyunjin and Lee, Kyoung Mu},
  journal = {arXiv preprint arXiv:2604.06989},
  year    = {2026},
  url     = {https://arxiv.org/abs/2604.06989}
}

@inproceedings{geng2024anagrams,
  author    = {Geng, Daniel and Park, Inbum and Owens, Andrew},
  title     = {Visual Anagrams: Generating Multi-View Optical Illusions
               with Diffusion Models},
  booktitle = {Proceedings of the IEEE/CVF Conference on Computer Vision
               and Pattern Recognition (CVPR)},
  year      = {2024}
}

@inproceedings{geng2024factorized,
  author    = {Geng, Daniel and Park, Inbum and Owens, Andrew},
  title     = {Factorized Diffusion: Perceptual Illusions by Noise
               Decomposition},
  booktitle = {European Conference on Computer Vision (ECCV)},
  year      = {2024}
}

@inproceedings{he2024scalecrafter,
  author    = {He, Yingqing and Yang, Shaoshu and Chen, Haoxin and Cun,
               Xiaodong and Xia, Menghan and Zhang, Yong and Wang, Xintao
               and He, Ran and Chen, Qifeng and Shan, Ying},
  title     = {{ScaleCrafter}: Tuning-free Higher-Resolution Visual
               Generation with Diffusion Models},
  booktitle = {International Conference on Learning Representations (ICLR)},
  year      = {2024}
}

@inproceedings{zhang2024hidiffusion,
  author    = {Zhang, Shen and Chen, Zhaowei and Zhao, Zhenyu and Chen,
               Yuhao and Tang, Yao and Liang, Jiajun},
  title     = {{HiDiffusion}: Unlocking Higher-Resolution Creativity and
               Efficiency in Pretrained Diffusion Models},
  booktitle = {European Conference on Computer Vision (ECCV)},
  pages     = {145--161},
  year      = {2024}
}

@inproceedings{huang2024fouriscale,
  author    = {Huang, Linjiang and Fang, Rongyao and Zhang, Aiping and
               Song, Guanglu and Liu, Si and Liu, Yu and Li, Hongsheng},
  title     = {{FouriScale}: A Frequency Perspective on Training-Free
               High-Resolution Image Synthesis},
  booktitle = {European Conference on Computer Vision (ECCV)},
  year      = {2024}
}

@article{ye2023ipadapter,
  title   = {{IP-Adapter}: Text Compatible Image Prompt Adapter for
             Text-to-Image Diffusion Models},
  author  = {Ye, Hu and Zhang, Jun and Liu, Sibo and Han, Xiao and
             Yang, Wei},
  journal = {arXiv preprint arXiv:2308.06721},
  year    = {2023},
  url     = {https://arxiv.org/abs/2308.06721}
}

@inproceedings{zhang2023controlnet,
  author    = {Zhang, Lvmin and Rao, Anyi and Agrawala, Maneesh},
  title     = {Adding Conditional Control to Text-to-Image Diffusion Models},
  booktitle = {Proceedings of the IEEE/CVF International Conference on
               Computer Vision (ICCV)},
  year      = {2023}
}

@article{labs2025fluxkontext,
  title   = {{FLUX.1} Kontext: Flow Matching for In-Context Image
             Generation and Editing in Latent Space},
  author  = {{Black Forest Labs} and Batifol, Stephen and Blattmann,
             Andreas and Boesel, Frederic and Consul, Saksham and
             Diagne, Cyril and Dockhorn, Tim and English, Jack and
             English, Zion and Esser, Patrick and Kulal, Sumith and
             Lacey, Kyle and Levi, Yam and Li, Cheng and Lorenz,
             Dominik and M{\"u}ller, Jonas and Podell, Dustin and
             Rombach, Robin and Saini, Harry and Sauer, Axel and
             Smith, Luke},
  journal = {arXiv preprint arXiv:2506.15742},
  year    = {2025},
  url     = {https://arxiv.org/abs/2506.15742}
}

@inproceedings{peebles2023dit,
  author    = {Peebles, William and Xie, Saining},
  title     = {Scalable Diffusion Models with Transformers},
  booktitle = {Proceedings of the IEEE/CVF International Conference on
               Computer Vision (ICCV)},
  year      = {2023}
}

@inproceedings{chen2024pixart,
  author    = {Chen, Junsong and Yu, Jincheng and Ge, Chongjian and Yao,
               Lewei and Xie, Enze and Wu, Yue and Wang, Zhongdao and Kwok,
               James and Luo, Ping and Lu, Huchuan and Li, Zhenguo},
  title     = {{PixArt-$\alpha$}: Fast Training of Diffusion Transformer for
               Photorealistic Text-to-Image Synthesis},
  booktitle = {International Conference on Learning Representations (ICLR)},
  year      = {2024}
}

@inproceedings{xie2025sana,
  author    = {Xie, Enze and Chen, Junsong and Chen, Junyu and Cai, Han and
               Tang, Haotian and Lin, Yujun and Zhang, Zhekai and Li, Muyang
               and Zhu, Ligeng and Lu, Yao and Han, Song},
  title     = {{SANA}: Efficient High-Resolution Image Synthesis with Linear
               Diffusion Transformers},
  booktitle = {International Conference on Learning Representations (ICLR)},
  year      = {2025}
}

@article{wu2025qwenimage,
  title   = {Qwen-Image Technical Report},
  author  = {Wu, Chenfei and Li, Jiahao and Zhou, Jingren and Lin, Junyang
             and Gao, Kaiyuan and Yan, Kun and Yin, Sheng-ming and Bai,
             Shuai and Xu, Xiao and Chen, Yilei and others},
  journal = {arXiv preprint arXiv:2508.02324},
  year    = {2025}
}

@article{zhao2026lens,
  title   = {{Lens}: Rethinking Training Efficiency for Foundational
             Text-to-Image Models},
  author  = {Guo, Baining and Luo, Chong and Chen, Dong and Chen,
             Dongdong and Wei, Fangyun and Li, Ji and Bao, Jianmin and
             Zhang, Jiawei and Zhao, Jinjing and Shi, Lei and Yang,
             Qinhong and Zhang, Sirui and Wu, Xiuyu and Feng, Xuelu and
             Lu, Yan and Dong, Yanchen and Yue, Yang and Wang, Yitong
             and Chen, Yunuo and Liang, Zhiyang and Wan, Ziyu},
  journal = {arXiv preprint arXiv:2605.21573},
  year    = {2026},
  url     = {https://arxiv.org/abs/2605.21573}
}

@misc{ideogram2026,
  author       = {{Ideogram}},
  title        = {{Ideogram 4.0}: An Open-Weight Text-to-Image Foundation
                  Model},
  year         = {2026},
  howpublished = {\url{https://huggingface.co/ideogram-ai}},
  note         = {Open-weight release, June 2026}
}

@article{rajabi2026sega,
  title   = {{SEGA}: Spectral-Energy Guided Attention for Resolution
             Extrapolation in Diffusion Transformers},
  author  = {Rajabi, Javad and Shaban, Kimia and Roohi, Koorosh and
             Lindell, David B and Taati, Babak},
  journal = {arXiv preprint arXiv:2605.22668},
  year    = {2026},
  url     = {https://arxiv.org/abs/2605.22668}
}

@inproceedings{koh2025scalediff,
  author    = {Koh, Sungho and Cha, SeungJu and Oh, Hyunwoo and Lee,
               Kwanyoung and Kim, Dong-Jin},
  title     = {{ScaleDiff}: Higher-Resolution Image Synthesis via Efficient
               and Model-Agnostic Diffusion},
  booktitle = {Advances in Neural Information Processing Systems (NeurIPS)},
  year      = {2025}
}

@inproceedings{qiu2024freescale,
  author    = {Qiu, Haonan and Zhang, Shiwei and Wei, Yujie and Chu, Ruihang
               and Yuan, Hangjie and Wang, Xiang and Zhang, Yingya and Liu,
               Ziwei},
  title     = {{FreeScale}: Unleashing the Resolution of Diffusion Models via
               Tuning-Free Scale Fusion},
  booktitle = {Proceedings of the IEEE/CVF International Conference on
               Computer Vision (ICCV)},
  year      = {2025}
}

@inproceedings{tragakis2024pixelsmith,
  author    = {Tragakis, Athanasios and Aversa, Marco and Kaul, Chaitanya
               and Murray-Smith, Roderick and Faccio, Daniele},
  title     = {Is One {GPU} Enough? Pushing Image Generation at
               Higher-Resolutions with Foundation Models},
  booktitle = {Advances in Neural Information Processing Systems (NeurIPS)},
  year      = {2024}
}

@inproceedings{kim2025diffusehigh,
  author    = {Kim, Younghyun and Hwang, Geunmin and Zhang, Junyu and Park,
               Eunbyung},
  title     = {{DiffuseHigh}: Training-Free Progressive High-Resolution Image
               Synthesis through Structure Guidance},
  booktitle = {Proceedings of the AAAI Conference on Artificial Intelligence},
  pages     = {4338--4346},
  year      = {2025}
}

@inproceedings{ciesielski2007,
  title        = {Evolution of Animated Photomosaics},
  author       = {Ciesielski, Vic and Berry, Marsha and Trist, Karen and
                  D'Souza, Daryl},
  booktitle    = {Workshops on Applications of Evolutionary Computation},
  pages        = {498--507},
  year         = {2007},
  organization = {Springer}
}

@article{he2019composing,
  title     = {Composing Photomosaic Images Using Clustering Based
               Evolutionary Programming},
  author    = {He, Yaodong and Zhou, Jianfeng and Yuen, Shiu Yin},
  journal   = {Multimedia Tools and Applications},
  volume    = {78},
  number    = {18},
  pages     = {25919--25936},
  year      = {2019},
  publisher = {Springer}
}

@inproceedings{rombach2022ldm,
  author    = {Rombach, Robin and Blattmann, Andreas and Lorenz, Dominik and
               Esser, Patrick and Ommer, Bj\"{o}rn},
  title     = {High-Resolution Image Synthesis with Latent Diffusion Models},
  booktitle = {Proceedings of the IEEE/CVF Conference on Computer Vision and
               Pattern Recognition (CVPR)},
  pages     = {10684--10695},
  year      = {2022}
}

@inproceedings{huang2017adain,
  author    = {Huang, Xun and Belongie, Serge},
  title     = {Arbitrary Style Transfer in Real-Time with Adaptive Instance
               Normalization},
  booktitle = {Proceedings of the IEEE/CVF International Conference on
               Computer Vision (ICCV)},
  pages     = {1501--1510},
  year      = {2017}
}

@article{mou2023t2i,
  title   = {{T2I-Adapter}: Learning Adapters to Dig out More Controllable
             Ability for Text-to-Image Diffusion Models},
  author  = {Mou, Chong and Wang, Xintao and Xie, Liangbin and Wu, Yanze
             and Zhang, Jian and Qi, Zhongang and Shan, Ying and Qie,
             Xiaohu},
  journal = {arXiv preprint arXiv:2302.08453},
  year    = {2023}
}

@article{kodaira2023streamdiffusion,
  title   = {{StreamDiffusion}: A Pipeline-Level Solution for Real-Time
             Interactive Generation},
  author  = {Kodaira, Akio and Xu, Chenfeng and Hazama, Toshiki and
             Yoshimoto, Takanori and Ohno, Kohei and Mitsuhori, Shogo and
             Sugano, Soichi and Cho, Hanying and Liu, Zhijian and Keutzer,
             Kurt},
  journal = {arXiv preprint arXiv:2312.12491},
  year    = {2023}
}

@inproceedings{wu2024hps,
  author    = {Wu, Xiaoshi and Hao, Yiming and Sun, Keqiang and Chen, Yixiong
               and Zhu, Feng and Zhao, Rui and Li, Hongsheng},
  title     = {Human Preference Score v2: A Solid Benchmark for Evaluating
               Human Preferences of Text-to-Image Synthesis},
  booktitle = {International Conference on Learning Representations (ICLR)},
  year      = {2024}
}

@inproceedings{xu2023imagereward,
  author    = {Xu, Jiazheng and Liu, Xiao and Wu, Yuchen and Tong, Yuxuan
               and Li, Qinghao and Ding, Ming and Tang, Jie and Dong, Yuxiao},
  title     = {{ImageReward}: Learning and Evaluating Human Preferences for
               Text-to-Image Generation},
  booktitle = {Advances in Neural Information Processing Systems (NeurIPS)},
  year      = {2023}
}

@inproceedings{radford2021clip,
  author    = {Radford, Alec and Kim, Jong Wook and Hallacy, Chris and
               Ramesh, Aditya and Goh, Gabriel and Agarwal, Sandhini and
               Sastry, Girish and Askell, Amanda and Mishkin, Pamela and
               Clark, Jack and Krueger, Gretchen and Sutskever, Ilya},
  title     = {Learning Transferable Visual Models from Natural Language
               Supervision},
  booktitle = {International Conference on Machine Learning (ICML)},
  pages     = {8748--8763},
  year      = {2021}
}

@inproceedings{li2022blip,
  author    = {Li, Junnan and Li, Dongxu and Xiong, Caiming and Hoi,
               Steven},
  title     = {{BLIP}: Bootstrapping Language-Image Pre-Training for Unified
               Vision-Language Understanding and Generation},
  booktitle = {International Conference on Machine Learning (ICML)},
  pages     = {12888--12900},
  year      = {2022}
}

@inproceedings{wang2023clipiqa,
  author    = {Wang, Jianyi and Chan, Kelvin C.K. and Loy, Chen Change},
  title     = {Exploring {CLIP} for Assessing the Look and Feel of Images},
  booktitle = {Proceedings of the AAAI Conference on Artificial Intelligence},
  pages     = {2555--2563},
  year      = {2023}
}

@inproceedings{brooks2023instructpix2pix,
  title     = {{InstructPix2Pix}: Learning to Follow Image Editing
               Instructions},
  author    = {Brooks, Tim and Holynski, Aleksander and Efros, Alexei A},
  booktitle = {Proceedings of the IEEE/CVF Conference on Computer Vision and
               Pattern Recognition (CVPR)},
  pages     = {18392--18402},
  year      = {2023}
}

@inproceedings{podell2024sdxl,
  title     = {{SDXL}: Improving Latent Diffusion Models for High-Resolution
               Image Synthesis},
  author    = {Podell, Dustin and English, Zion and Lacey, Kyle and
               Blattmann, Andreas and Dockhorn, Tim and M{\"u}ller, Jonas and
               Penna, Joe and Rombach, Robin},
  booktitle = {International Conference on Learning Representations (ICLR)},
  year      = {2024}
}

@inproceedings{doyle2026diffusion,
  author    = {Doyle, Lars and Mould, David},
  title     = {Diffusion-based Image Mosaics},
  year      = {2026},
  publisher = {Association for Computing Machinery},
  address   = {New York, NY, USA},
  booktitle = {Proceedings of Graphics Interface},
  series    = {GI '26},
  numpages  = {10}
}

@article{ho2020ddpm,
  title   = {Denoising Diffusion Probabilistic Models},
  author  = {Ho, Jonathan and Jain, Ajay and Abbeel, Pieter},
  journal = {Advances in Neural Information Processing Systems},
  volume  = {33},
  pages   = {6840--6851},
  year    = {2020}
}

@article{song2020score,
  title   = {Score-Based Generative Modeling through Stochastic Differential
             Equations},
  author  = {Song, Yang and Sohl-Dickstein, Jascha and Kingma, Diederik P and
             Kumar, Abhishek and Ermon, Stefano and Poole, Ben},
  journal = {arXiv preprint arXiv:2011.13456},
  year    = {2020}
}

@inproceedings{zhang2025diffusion,
  title     = {Diffusion-4k: Ultra-High-Resolution Image Synthesis with
               Latent Diffusion Models},
  author    = {Zhang, Jinjin and Huang, Qiuyu and Liu, Junjie and Guo,
               Xiefan and Huang, Di},
  booktitle = {Proceedings of the Computer Vision and Pattern Recognition
               Conference},
  pages     = {23464--23473},
  year      = {2025}
}

@article{wang2004image,
  title     = {Image Quality Assessment: From Error Visibility to Structural
               Similarity},
  author    = {Wang, Zhou and Bovik, Alan C and Sheikh, Hamid R and
               Simoncelli, Eero P},
  journal   = {IEEE Transactions on Image Processing},
  volume    = {13},
  number    = {4},
  pages     = {600--612},
  year      = {2004},
  publisher = {IEEE}
}

@inproceedings{zhang2018unreasonable,
  title     = {The Unreasonable Effectiveness of Deep Features as a
               Perceptual Metric},
  author    = {Zhang, Richard and Isola, Phillip and Efros, Alexei A and
               Shechtman, Eli and Wang, Oliver},
  booktitle = {Proceedings of the IEEE Conference on Computer Vision and
               Pattern Recognition (CVPR)},
  pages     = {586--595},
  year      = {2018}
}

@inproceedings{lee2024diffusion,
  title        = {Diffusion-Based Image-to-Image Translation by Noise
                  Correction via Prompt Interpolation},
  author       = {Lee, Junsung and Kang, Minsoo and Han, Bohyung},
  booktitle    = {European Conference on Computer Vision (ECCV)},
  pages        = {289--304},
  year         = {2024},
  organization = {Springer}
}

@inproceedings{shum2025color,
  title     = {Color Alignment in Diffusion},
  author    = {Shum, Ka Chun and Hua, Binh-Son and Nguyen, Duc Thanh and
               Yeung, Sai-Kit},
  booktitle = {Proceedings of the IEEE/CVF Conference on Computer Vision and
               Pattern Recognition (CVPR)},
  pages     = {28446--28455},
  year      = {2025}
}

@inproceedings{hajiali2024elasticdiffusion,
  title     = {ElasticDiffusion: Training-free Arbitrary Size Image Generation
               through Global-Local Content Separation},
  author    = {Haji-Ali, Moayed and Balakrishnan, Guha and Ordonez, Vicente},
  booktitle = {Proceedings of the IEEE/CVF Conference on Computer Vision and
               Pattern Recognition (CVPR)},
  year      = {2024}
}

@inproceedings{frolov2025spotdiffusion,
  title     = {SpotDiffusion: A Fast Approach for Seamless Panorama Generation
               Over Time},
  author    = {Frolov, Stanislav and Moser, Brian B. and Dengel, Andreas},
  booktitle = {Proceedings of the IEEE/CVF Winter Conference on Applications of
               Computer Vision (WACV)},
  year      = {2025}
}

@inproceedings{lee2023syncdiffusion,
  title     = {SyncDiffusion: Coherent Montage via Synchronized Joint Diffusions},
  author    = {Lee, Yuseung and Kim, Kunho and Kim, Hyunjin and Sung, Minhyuk},
  booktitle = {Advances in Neural Information Processing Systems (NeurIPS)},
  year      = {2023}
}

@misc{fid,
      title={GANs Trained by a Two Time-Scale Update Rule Converge to a Local Nash Equilibrium}, 
      author={Martin Heusel and Hubert Ramsauer and Thomas Unterthiner and Bernhard Nessler and Sepp Hochreiter},
      year={2018},
      eprint={1706.08500},
      archivePrefix={arXiv},
      primaryClass={cs.LG},
      url={https://arxiv.org/abs/1706.08500}, 
}
}

\clearpage
\appendix

\begin{table*}[t]
\centering
\caption{Global structure preservation across downsampling scales, mean (std) over all samples. The $64\times64$ scale reported in the main paper is omitted here. All methods use SD2.1 with identical base image, prompts, resolution and grid. HPS: HPSv2, IR: Image Reward. \textbf{Bold} denotes the best result and \underline{underline} the second best in each column.}
\label{tab:multiscale}
\resizebox{\textwidth}{!}{
\scriptsize
\begin{tabular}{l ccccc ccccc}
\toprule
\multirow{2}{*}{\textbf{Method}} & \multicolumn{5}{c}{\textbf{PSNR $\uparrow$}} & \multicolumn{5}{c}{\textbf{SSIM $\uparrow$}} \\
\cmidrule(lr){2-6} \cmidrule(lr){7-11}
& $32^2$ & $128^2$ & $256^2$ & $512^2$ & $768^2$ & $32^2$ & $128^2$ & $256^2$ & $512^2$ & $768^2$ \\
\midrule
Match \& Tone & 17.39\sd{2.12} & 14.68\sd{2.14} & 14.08\sd{2.19} & 13.69\sd{2.24} & 13.54\sd{2.26} & 0.49\sd{0.11} & 0.21\sd{0.11} & 0.22\sd{0.13} & 0.27\sd{0.14} & 0.30\sd{0.15} \\
AdaIN & 22.27\sd{1.90} & 18.18\sd{2.05} & 17.23\sd{2.23} & 16.65\sd{2.36} & 16.44\sd{2.41} & 0.80\sd{0.07} & 0.42\sd{0.07} & 0.36\sd{0.09} & 0.37\sd{0.12} & 0.40\sd{0.13} \\
Color T2I-Adapter & 11.62\sd{1.36} & 10.48\sd{1.10} & 10.13\sd{1.05} & 9.89\sd{1.02} & 9.80\sd{1.02} & 0.18\sd{0.06} & 0.11\sd{0.03} & 0.10\sd{0.04} & 0.12\sd{0.07} & 0.15\sd{0.08} \\
NoiseBlend & 15.42\sd{1.78} & 14.24\sd{1.85} & 14.01\sd{1.87} & 13.86\sd{1.89} & 13.80\sd{1.90} & 0.41\sd{0.12} & 0.28\sd{0.10} & 0.30\sd{0.11} & 0.35\sd{0.12} & 0.39\sd{0.12} \\
StreamDiff & \underline{28.08}\sd{4.39} & \underline{24.80}\sd{2.96} & \underline{23.30}\sd{2.65} & \underline{21.97}\sd{2.66} & \underline{21.49}\sd{2.71} & \textbf{0.94}\sd{0.05} & \underline{0.81}\sd{0.05} & \underline{0.72}\sd{0.05} & \underline{0.64}\sd{0.07} & \textbf{0.62}\sd{0.08} \\
Phomosaic & 22.89\sd{1.43} & 19.09\sd{1.47} & 18.04\sd{1.52} & 17.49\sd{1.57} & 17.30\sd{1.60} & 0.78\sd{0.05} & 0.42\sd{0.04} & 0.34\sd{0.05} & 0.35\sd{0.08} & 0.38\sd{0.09} \\
\midrule
PhotoQuilt (SD2.1) & \textbf{30.05}\sd{1.11} & \textbf{25.09}\sd{1.43} & \textbf{24.84}\sd{1.57} & \textbf{23.20}\sd{1.68} & \textbf{21.96}\sd{1.73} & \underline{0.92}\sd{0.03} & \textbf{0.82}\sd{0.05} & \textbf{0.80}\sd{0.05} & \textbf{0.76}\sd{0.07} & \underline{0.58}\sd{0.08} \\
\midrule
\multirow{2}{*}{\textbf{Method}} & \multicolumn{5}{c}{\textbf{LPIPS $\downarrow$}} & \multicolumn{5}{c}{\textbf{HPS $\uparrow$}} \\
\cmidrule(lr){2-6} \cmidrule(lr){7-11}
& $32^2$ & $128^2$ & $256^2$ & $512^2$ & $768^2$ & $32^2$ & $128^2$ & $256^2$ & $512^2$ & $768^2$ \\
\midrule
Match \& Tone & 0.075\sd{0.037} & 0.519\sd{0.089} & 0.569\sd{0.084} & 0.626\sd{0.074} & 0.651\sd{0.068} & 0.124\sd{0.021} & 0.124\sd{0.022} & 0.140\sd{0.022} & 0.143\sd{0.022} & 0.143\sd{0.022} \\
AdaIN & 0.023\sd{0.017} & 0.433\sd{0.082} & \underline{0.535}\sd{0.069} & 0.599\sd{0.056} & 0.619\sd{0.054} & 0.159\sd{0.029} & 0.139\sd{0.025} & 0.160\sd{0.026} & 0.162\sd{0.026} & 0.163\sd{0.026} \\
Color T2I-Adapter & 0.301\sd{0.091} & 0.651\sd{0.077} & 0.670\sd{0.054} & 0.725\sd{0.051} & 0.757\sd{0.049} & 0.114\sd{0.022} & 0.129\sd{0.020} & 0.149\sd{0.020} & 0.151\sd{0.020} & 0.151\sd{0.020} \\
NoiseBlend & 0.109\sd{0.043} & 0.445\sd{0.092} & 0.595\sd{0.104} & 0.717\sd{0.100} & 0.743\sd{0.098} & 0.195\sd{0.030} & 0.218\sd{0.032} & 0.220\sd{0.032} & 0.219\sd{0.032} & 0.218\sd{0.032} \\
StreamDiff & \underline{0.006}\sd{0.007} & \textbf{0.079}\sd{0.047} & \textbf{0.164}\sd{0.068} & \textbf{0.286}\sd{0.075} & \textbf{0.342}\sd{0.069} & \underline{0.221}\sd{0.028} & \underline{0.250}\sd{0.039} & \underline{0.272}\sd{0.041} & \textbf{0.277}\sd{0.042} & \textbf{0.277}\sd{0.042} \\
Phomosaic & 0.026\sd{0.011} & 0.427\sd{0.064} & 0.573\sd{0.068} & 0.684\sd{0.067} & 0.746\sd{0.078} & 0.172\sd{0.021} & 0.154\sd{0.024} & 0.171\sd{0.024} & 0.174\sd{0.025} & 0.174\sd{0.025} \\
\midrule
PhotoQuilt (SD2.1) & \textbf{0.005}\sd{0.005} & \underline{0.132}\sd{0.043} & \textbf{0.164}\sd{0.059} & \underline{0.402}\sd{0.064} & \underline{0.486}\sd{0.076} & \textbf{0.222}\sd{0.022} & \textbf{0.263}\sd{0.030} & \textbf{0.273}\sd{0.027} & \underline{0.275}\sd{0.027} & \underline{0.275}\sd{0.028} \\
\midrule
\multirow{2}{*}{\textbf{Method}} & \multicolumn{5}{c}{\textbf{IR $\uparrow$}} & \multicolumn{5}{c}{\textbf{FID $\downarrow$}} \\
\cmidrule(lr){2-6} \cmidrule(lr){7-11}
& $32^2$ & $128^2$ & $256^2$ & $512^2$ & $768^2$ & $32^2$ & $128^2$ & $256^2$ & $512^2$ & $768^2$ \\
\midrule
Match \& Tone & -2.27\sd{0.03} & -2.26\sd{0.06} & -2.24\sd{0.09} & -2.23\sd{0.10} & -2.23\sd{0.10} & 202.15 & 203.51 & 212.19 & 213.41 & 213.54 \\
AdaIN & -2.13\sd{0.31} & -2.11\sd{0.37} & -2.08\sd{0.41} & -2.07\sd{0.42} & -2.07\sd{0.42} & 157.37 & 215.37 & 226.14 & 228.81 & 229.04 \\
Color T2I-Adapter & -2.28\sd{0.03} & -2.26\sd{0.05} & -2.25\sd{0.07} & -2.25\sd{0.07} & -2.25\sd{0.07} & 308.23 & 303.63 & 293.87 & 288.53 & 288.30 \\
NoiseBlend & -1.29\sd{0.90} & -0.10\sd{1.04} & -0.08\sd{1.04} & -0.08\sd{1.03} & -0.08\sd{1.03} & \textbf{118.32} & \underline{137.39} & \underline{153.61} & \textbf{157.74} & \textbf{158.38} \\
StreamDiff & \underline{-1.06}\sd{1.03} & \textbf{0.72}\sd{0.96} & \textbf{0.88}\sd{0.93} & \underline{0.90}\sd{0.92} & \underline{0.90}\sd{0.92} & \underline{122.64} & 154.07 & 163.20 & \underline{164.42} & \underline{164.51} \\
Phomosaic & -2.10\sd{0.33} & -2.17\sd{0.19} & -2.16\sd{0.19} & -2.15\sd{0.21} & -2.14\sd{0.21} & 179.04 & 260.90 & 293.66 & 303.21 & 303.67 \\
\midrule
PhotoQuilt (SD2.1) & \textbf{-0.93}\sd{0.80} & \underline{0.60}\sd{0.84} & \underline{0.74}\sd{0.87} & \textbf{0.98}\sd{0.86} & \textbf{0.98}\sd{0.86} & 122.74 & \textbf{135.21} & \textbf{148.35} & 172.32 & 172.85 \\
\bottomrule
\end{tabular}
}
\end{table*}

\section{Implementation Details}
All experiments were executed on NVIDIA H100 GPUs. For our generative backbones, we used the following specific model checkpoints: \texttt{FLUX.1-Krea-dev-12B} for the FLUX.1~\cite{flux2024} evaluations, \texttt{FLUX.2-Klein-9B} for FLUX.2~\cite{flux2_2025}, and \texttt{stable-diffusion-2-1-base} for Stable Diffusion 2.1~\cite{rombach2022ldm}. Prompts were selected to span scene type (landscape, portrait, architecture, still life), dominant object scale, colour range, and low-frequency contrast, since the last determines how tightly the shared latent constrains each tile.

All baselines were run with their authors' released configurations and default hyperparameters; no method-specific tuning was performed for any method, including ours. Where official source code for baseline methods was unavailable, we faithfully re-implemented the algorithms according to the technical specifications provided in their original publications. Every method received an identical base image, an identical global prompt, identical tile prompts, an identical target resolution and an identical partition. For the same backbone comparison, all baseline methods used SD2.1 as well PhotoQuilt for a fair comparison.

\textbf{Match \& Tone.} Each target cell is matched against a fixed image pool by RGB block statistics, tone-adjusted toward the cell statistics and pasted. Tiles are generated using SD2.1 backbone. Tile reuse is permitted, so a source image may appear in several cells. Conditioning is limited to the block colour statistics.

\textbf{AdaIN.} Tiles are generated independently from the tile prompt. The predicted clean latent is matched per channel toward the mean and standard deviation of the corresponding target block latent at partial strength ($0.25$), giving colour and tone but no spatial structure.

\textbf{Color T2I-Adapter.} Each target block is reduced to an $8\times8$ colour palette, upsampled to tile resolution and supplied to the adapter as a spatial conditioning signal (conditioning scale $1.0$). The conditioning therefore carries no detail below the palette scale.

\textbf{NoiseBlend.} Two branches are formed per tile, one conditioned on the full target and global prompt and one on the target crop and tile prompt, combined at a fixed $0.5/0.5$ ratio. No renoising step is applied.

\textbf{StreamDiff.} All tiles are processed as a single batch, each initialised image-to-image from its own crop of the target and refined under the tile prompt with a short step schedule. No information is exchanged between tiles. Of all methods compared, this supplies each tile with the most direct form of its target region.

\textbf{Phomosaic.} A reference image is generated at block resolution; tiles are then generated independently and placed edge-to-edge with no overlap. Three components link each tile to its corresponding reference block: a single low-resolution Gaussian field, enlarged via integral upsampling, so adjacent tiles begin from correlated signals instead of pure noise; AdaIN matching of the clean estimate to the block latent statistics; and a block-scale low-frequency term pulling the clean estimate toward the reference block ($w{=}0.7$) for the first $85\%$ of steps, after which tiles denoise freely.

\begin{figure*}[pt!]
  \centering
  \includegraphics[width=\textwidth]{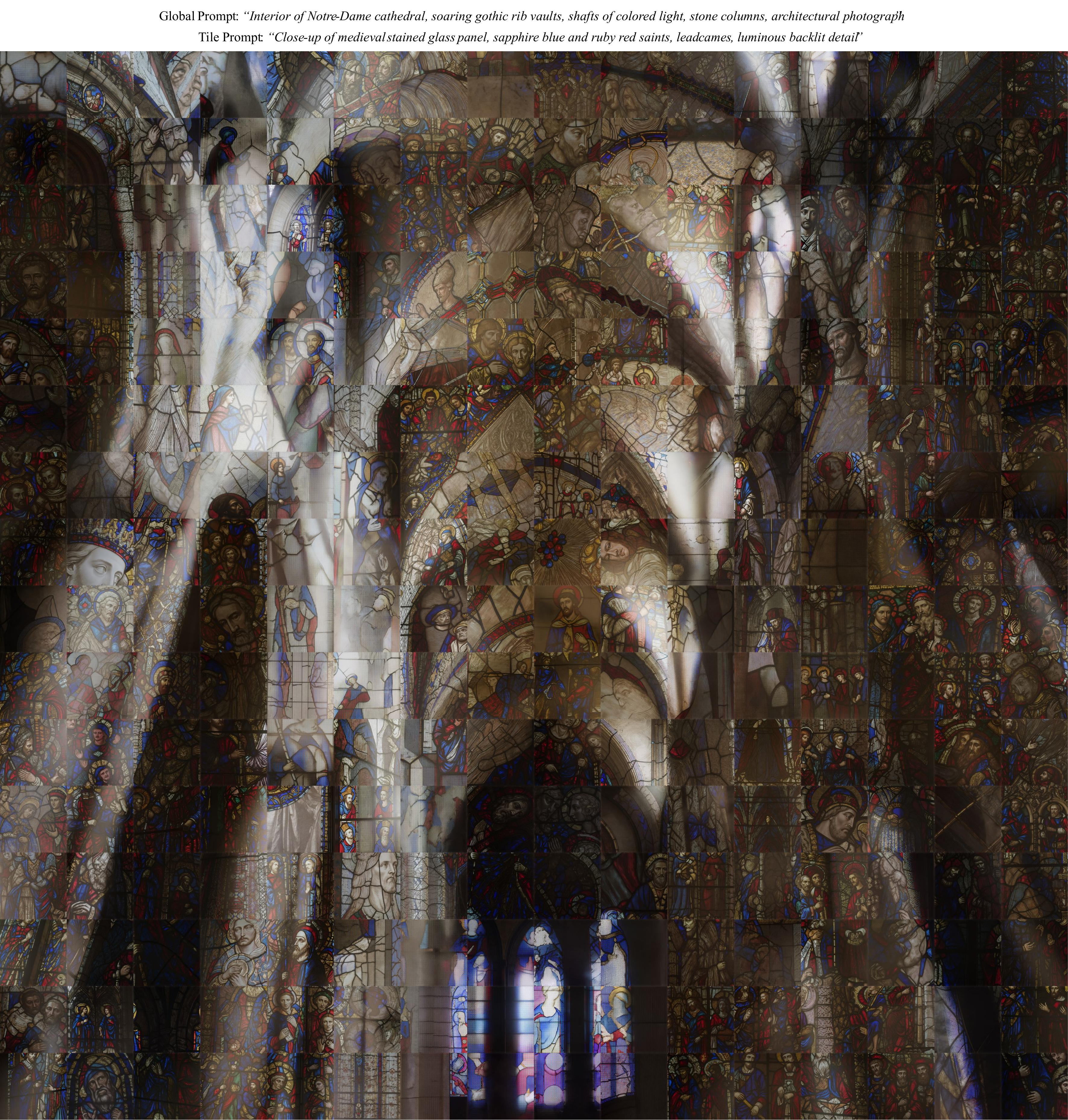}
  \caption{Photomosaic (8192 x 8192) generated with text prompts while the text prompt for the tiles (512 x 512) is different than the global image.}
  \label{fig:8k-sample}
\end{figure*}

\section{Tile and Base Conditioning}
The dual-scale formulation of PhotoQuilt exposes a highly flexible conditioning interface with two independent axes: the global base condition and the local tile conditions. This allows the framework to operate in several distinct generative modes without altering the underlying bootstrapped denoising procedure.

\paragraph{Base (Global Structure).}
The global structure of the photomosaic is established in the first stage and can be driven by either text or vision. When generating a purely novel scene, the base is conditioned on a global text prompt $c_0$. Alternatively, to reconstruct a specific real-world target, the base can be initialized from an arbitrary reference image $I_0$. In this latter case, $I_0$ is encoded directly into the latent space using the backbone's specific encoder to serve as the structural anchor, ensuring the low-frequency layout faithfully matches the real target.

\begin{figure*}[pt!]
  \centering
  \includegraphics[width=0.61\textwidth]{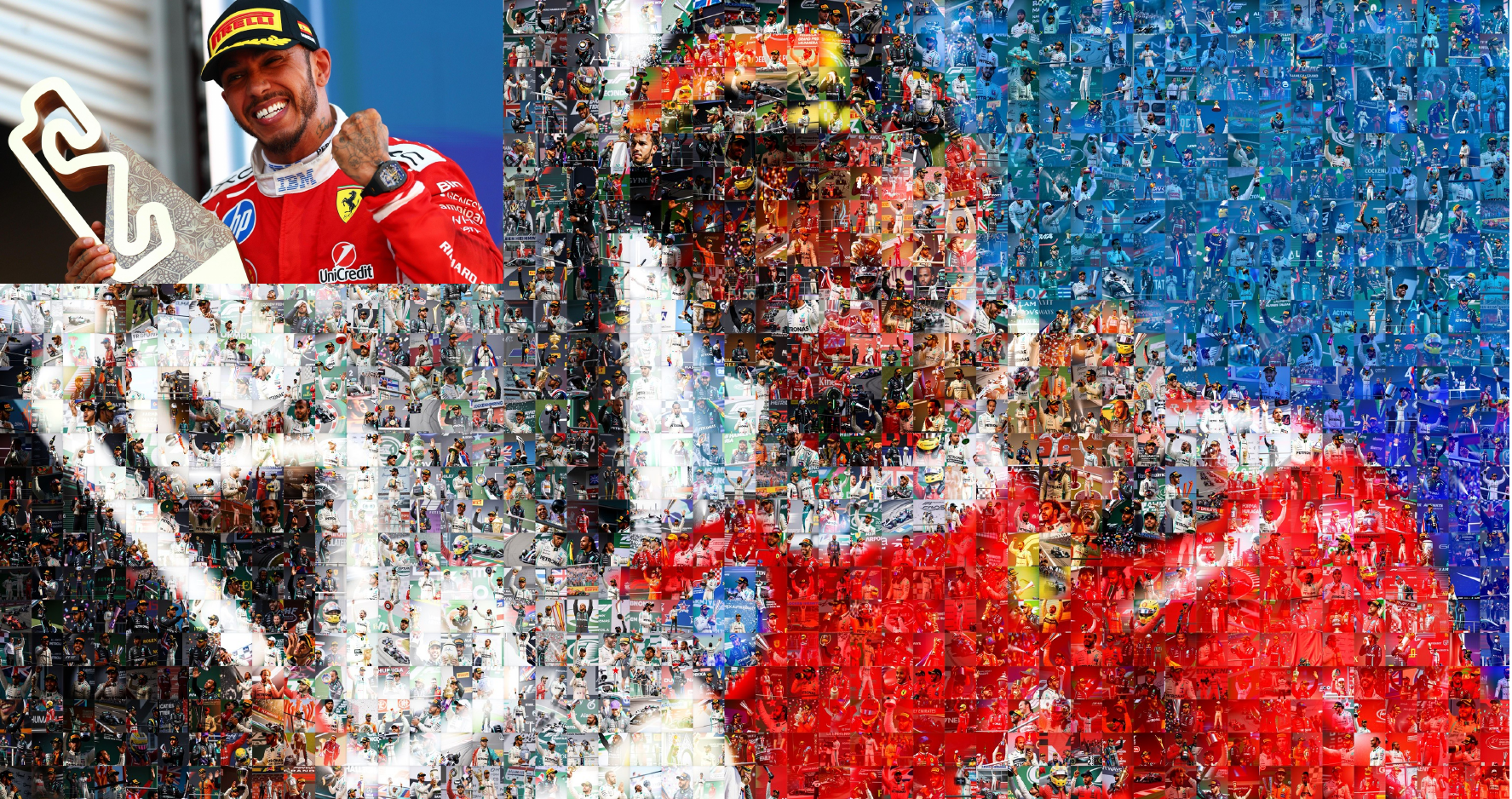}
  \caption{Photomosaic (12288 x 6144) generated from a base image (4096 x 2048) using real images as tiles (256 x 256) through image gallery conditioning. The gallery has been fetched from web and the base image has been encoded then upscaled to the preferred resolution.}
  \label{fig:hamilton-full}
\end{figure*}

\paragraph{Tiles (Local Content).}
During the independent per-tile denoising phase, each tile is guided by its own condition $c_k$, has its own separated condition tokens positioned at each tile's origin point. PhotoQuilt supports three primary modes for tile-level conditioning:
\begin{enumerate}
    \item \textbf{Shared Global Prompt:} By default, $c_k = c_0$. Even when all tiles share the same global text condition, the independent denoising trajectories and the restricted attention mask force the tiles to diverge into distinct, self-contained images that collectively respect the target layout.
    \item \textbf{Per-Tile Text Prompts:} For fine-grained semantic control, each tile can be guided by a unique text prompt $c_k^{\mathrm{tile}}$, allowing explicit control over the subject matter of individual tiles. A high resolution sample of different prompts for global image and tiles is shown in Fig.~\ref{fig:8k-sample}.
    \item \textbf{Image Gallery Conditioning:} To replicate the classical retrieval-based photomosaic experience using generative AI, tiles can be conditioned on real images drawn from a provided gallery $\mathcal{G}=\{g_1,\dots,g_M\}$. We assign a gallery entry to each tile via a mapping function $\pi:\{1,\dots,K\}\!\to\!\{1,\dots,M\}$ that samples uniformly at random from the gallery. The tile is then conditioned through the backbone's native image interface,
    \begin{equation}
      c_k = \mathcal{R}\big(g_{\pi(k)}\big),
      \label{eq:gallery}
    \end{equation}
    where $\mathcal{R}$ represents the image conditioning adapter (Redux~\cite{fluxredux2024} for FLUX.1, and the built-in image-to-image conditioning for FLUX.2). This recovers the classical tile-pool setting, but uses each reference as generative guidance rather than a pasted patch. Consequently, the synthesized tiles adopt the semantic and stylistic characteristics of the reference images while structurally adapting to fit their region of the shared base latent. An example in high resolution is shown in Fig.~\ref{fig:hamilton-full}.
\end{enumerate}

\begin{table*}[t]
\centering
\caption{Inference time and peak GPU memory across grid configurations.}
\label{tab:performance_comparison}
\resizebox{\textwidth}{!}{
\scriptsize
\begin{tabular}{l cc cc cc}
\toprule
\multirow{2}{*}{\textbf{Method}}
& \multicolumn{2}{c}{\textbf{$8\times8$ grid, $768^2$ tiles, $6144^2$}}
& \multicolumn{2}{c}{\textbf{$6\times6$ grid, $768^2$ tiles, $4608^2$}}
& \multicolumn{2}{c}{\textbf{$12\times12$ grid, $512^2$ tiles, $6144^2$}} \\
\cmidrule(lr){2-3} \cmidrule(lr){4-5} \cmidrule(lr){6-7}
& Time (s) $\downarrow$ & Mem. (GB) $\downarrow$
& Time (s) $\downarrow$ & Mem. (GB) $\downarrow$
& Time (s) $\downarrow$ & Mem. (GB) $\downarrow$ \\
\midrule
Match \& Tone       & 61.09  & 5.2  & 59.87 & 5.2 & 116.82 & 4.9 \\
AdaIN               & 284.17 & 3.8  & 263.21 & 3.4 & 398.61 & 4.5 \\
Color T2I-Adapter   & 527.34 & 3.5  & 510.44 & 3.5 & 628.49 & 4.1 \\
NoiseBlend          & 346.74 & 3.8  & 340.51 & 3.5 & 393.11 & 4.2 \\
StreamDiff          & 87.57  & 5.1  & 80.65 & 4.5 & 146.33 & 5.5 \\
Phomosaic           & 267.22 & 3.8  & 191.98 & 3.5 & 388.49 & 4.5 \\
\midrule
PhotoQuilt (SD2.1)  & 97.15  & 4.2  & 95.57 & 4.0 & 145.97 & 4.2 \\
PhotoQuilt (FLUX.1) & 209.59 & 50.9 & 146.38 & 42.4 & 224.64 & 51.8 \\
PhotoQuilt (FLUX.2) & 139.41 & 68.6 & 104.17 & 52.7 & 154.76 & 69.7 \\
\bottomrule
\end{tabular}
}
\end{table*}

\section{Further Global Structure Evaluation}
\begin{figure*}[pt!]
  \centering
  \includegraphics[width=0.94\textwidth]{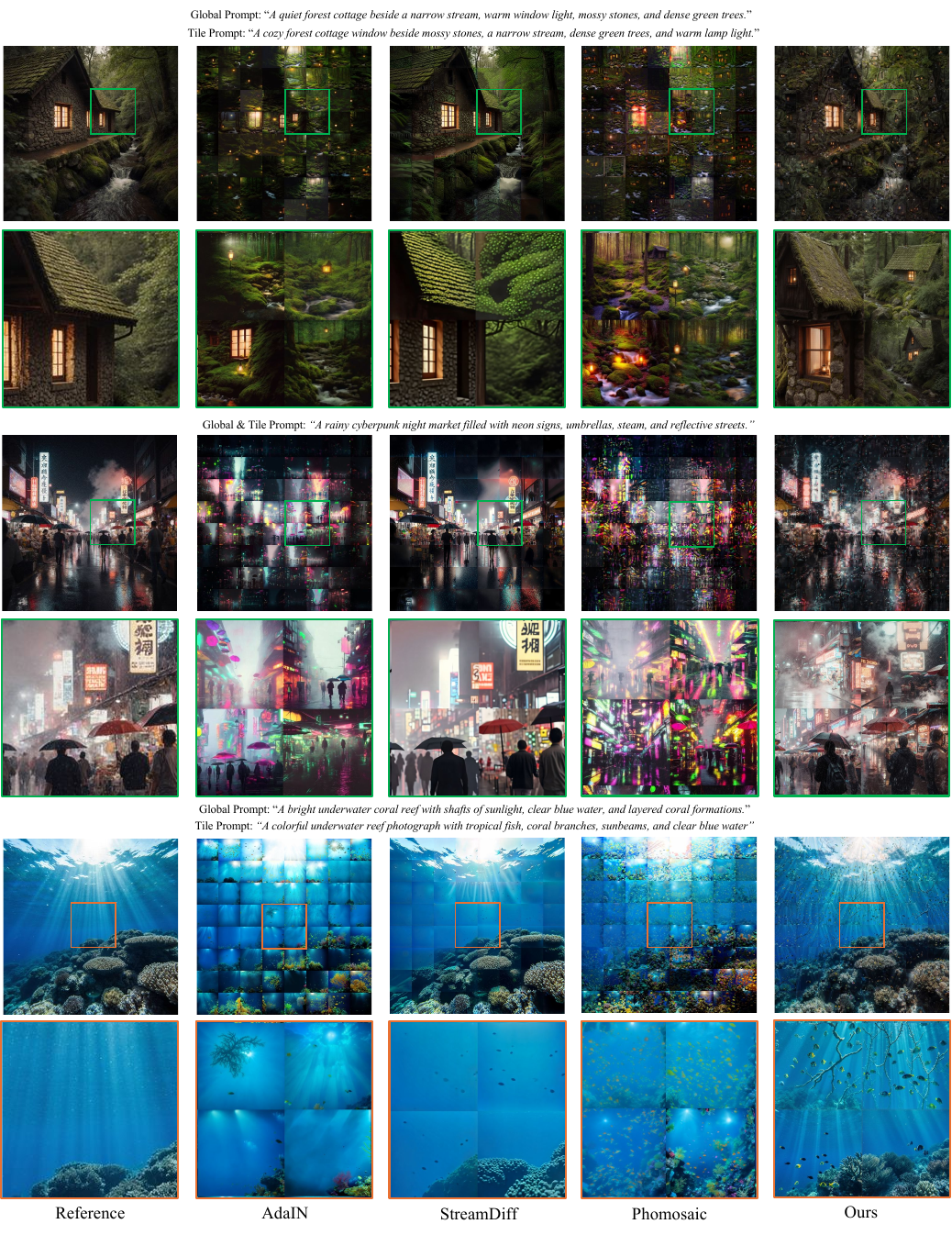}
  \caption{Further qualitative comparison of our method with the baselines across more diverse scenes.}
  \label{fig:qual-comp-alt}
\end{figure*}

Table~\ref{tab:multiscale} extends our global fidelity analysis to $32\times32$, $128\times128$, $256\times256$, $512\times512$, and $768\times768$ downsampling scales, capturing both coarse layout fidelity and mid-frequency structure preservation. While StreamDiff~\cite{kodaira2023streamdiffusion} posts strong pixel-level scores across all resolutions, this largely stems from its failure to produce independent tiles, collapsing the mosaic into a single continuous image. True mosaic baselines (Phomosaic~\cite{chung2026photomosaic}, AdaIN~\cite{huang2017adain}, Match \& Tone~\cite{silvers1997,finkelstein1998}) show significant structural degradation at finer scales ($128\times128$ and $256\times256$), confirming that mere color transfer cannot enforce rigid spatial alignment. PhotoQuilt demonstrates robust structural preservation across most of evaluated resolutions, while being comparable on the rest. More qualitative comparison is shown in Fig.~\ref{fig:qual-comp-alt}.

\begin{figure*}[pt!]
  \centering
  \includegraphics[width=\textwidth]{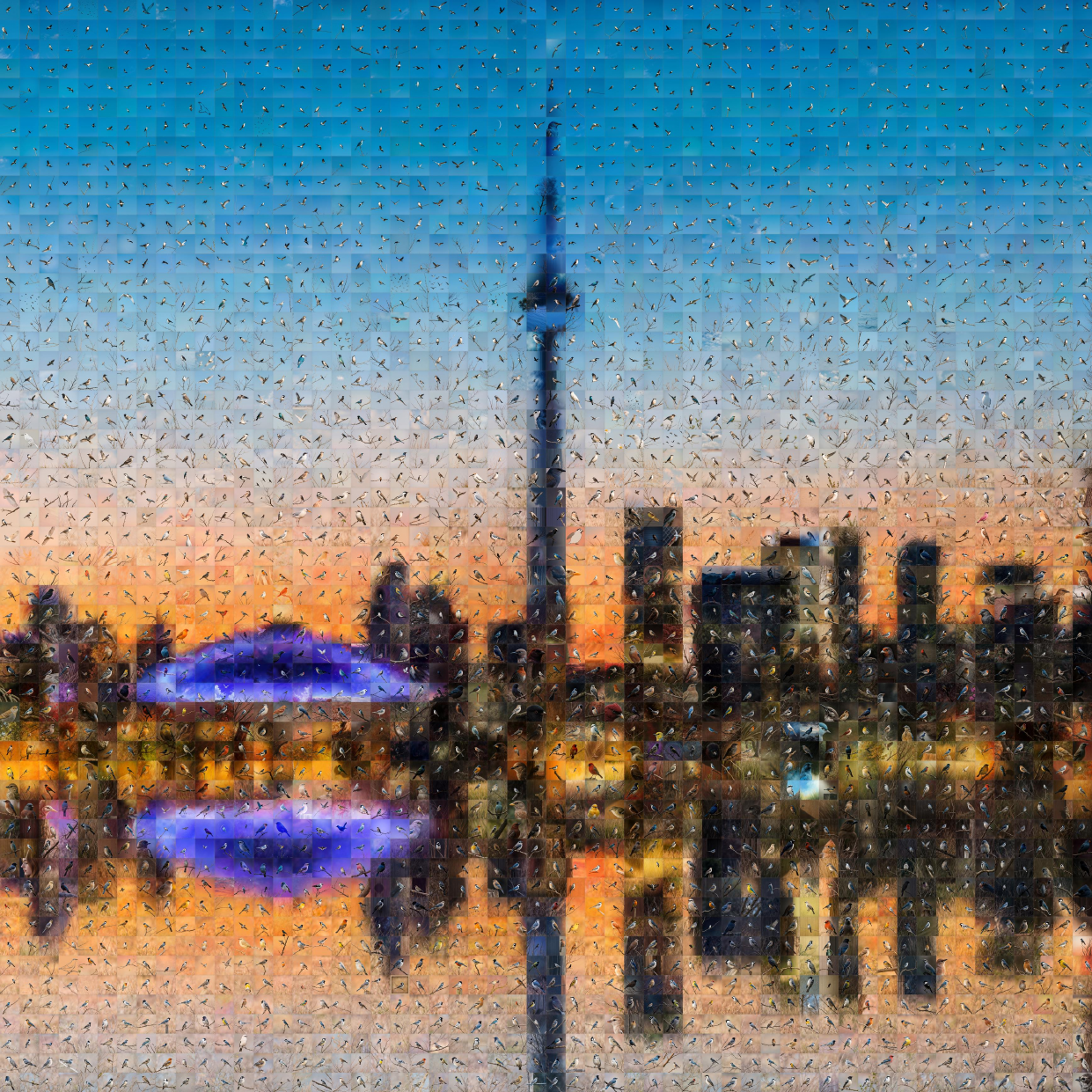}
  \caption{Photomosaic (14336 x 14336) generated using the multi-gpu distribution feature with 256 x 256 tiles. A base image has been used, and the text prompt for tiles was "A bird". The generation has been done using 4xH100 GPUs on a single node. Best viewed zoomed in.}
  \label{fig:dist-bird}
\end{figure*}

\section{Inference Time and Memory Analysis}

Table~\ref{tab:performance_comparison} details the inference times and peak GPU memory of PhotoQuilt and the evaluated baselines, measured at three grid configurations. While Match \& Tone reports the lowest inference time, this figure exclusively represents the matching and tone-adjustment phase; as a retrieval approach, it relies on a pre-computed pool of images generated, the substantial computational cost of which is excluded from this measurement. Despite this, PhotoQuilt operating on the SD2.1 backbone achieves speeds highly competitive with this retrieval lower bound. StreamDiff similarly posts fast execution times using the SD2.1 backbone by processing tiles in an uncoordinated batch. However, as established in our main evaluation, this high-throughput pipeline critically compromises mosaic quality by collapsing tile diversity. PhotoQuilt achieves comparable generation speeds on the same backbone while strictly preserving both global layout fidelity and local tile autonomy.

The algorithmic efficiency of our bootstrapped tiled denoising approach becomes most apparent when compared to Phomosaic, our primary fully generative competitor. As shown in Table~\ref{tab:performance_comparison}, when operating on the identical SD2.1 backbone, PhotoQuilt significantly outpaces Phomosaic at every configuration, demonstrating that our per-tile denoising mechanism is fundamentally faster than relying on iterative alignment losses and separate coordination steps. More notably, this architectural efficiency allows our method to comfortably scale to state-of-the-art architectures: PhotoQuilt remains faster than Phomosaic's SD2.1 implementation even when executing on the heavier FLUX.1 and FLUX.2 DiT backbones.

Across configurations, cost is governed by the total tile-pixel count together with a fixed per-run overhead, and contains no term that grows quadratically with canvas area, since attention is confined within each tile. At these canvas sizes the fixed overhead is substantial, so measured time grows sublinearly across our configurations: on SD2.1, generating $144$ tiles at $12\times12$ takes only $1.5\times$ the time of $64$ tiles at $8\times8$ on the same canvas, and the $6\times6$ configuration is not appreciably faster then $8\times8$. This follows directly from tile independence, which allows tiles to be denoised as a single parallel batch rather than sequentially. Peak memory correspondingly reflects the size of that batch rather than the canvas alone, and we do not claim it to be constant in output resolution. Because tiles share no state, however, the batch can be reduced to trade throughput for memory, or partitioned across devices with no communication between them. The latter is what allows PhotoQuilt to reach resolutions beyond the capacity of a single accelerator, as described in Sec.~\ref{sec:multigpu}. Peak memory for the FLUX rows is dominated by model weights rather than by mosaic construction; on the same backbone, PhotoQuilt requires memory comparable to the SD2.1 baselines.

\section{Multi-GPU Distributed Generation for Ultra-High-Resolution Photomosaics}
\label{sec:multigpu}

A key advantage of PhotoQuilt's tile-level denoising is its natural compatibility with distributed generation across multiple GPUs. This enables the synthesis of ultra-high-resolution canvases (e.g., 14k$\times$14k) without compromising quality or requiring approximation (see Fig.~\ref{fig:dist-bird}). Unlike standard global-attention diffusion models where every token attends to the entire image, our method confines generation to fixed-size, spatially local attention windows. This allows us to partition the upscaled global latent into horizontal bands split cleanly along tile-row boundaries. Each band is assigned to a separate GPU holding a full model replica, allowing denoising to proceed in parallel with equal load balancing. To guarantee that this distributed execution perfectly matches a single-GPU run, we enforce trajectory consistency: the timestep shift ($\mu$) in the FLUX denoising schedule is computed once based on the full-canvas sequence length and shared across all replicas. This prevents individual bands from calculating a localized shift and drifting onto divergent denoising paths.

Beyond denoising, the final pixel-space decoding via the Variational Autoencoder (VAE) presents a severe memory bottleneck; natively decoding even a full-width strip of a massive canvas easily exceeds standard GPU memory capacities. To make peak memory independent of the total canvas size, PhotoQuilt employs a two-dimensional, distributed block-tiling strategy for the decode phase. The latent is divided into small two-dimensional blocks, and these blocks are distributed round-robin across the GPUs and decoded. This yields an ultra-high-resolution output whose maximum scale is bounded only by aggregate host memory, transforming generation into a purely throughput-limited process.

\begin{figure*}[pt!]
  \centering
  \includegraphics[width=\textwidth]{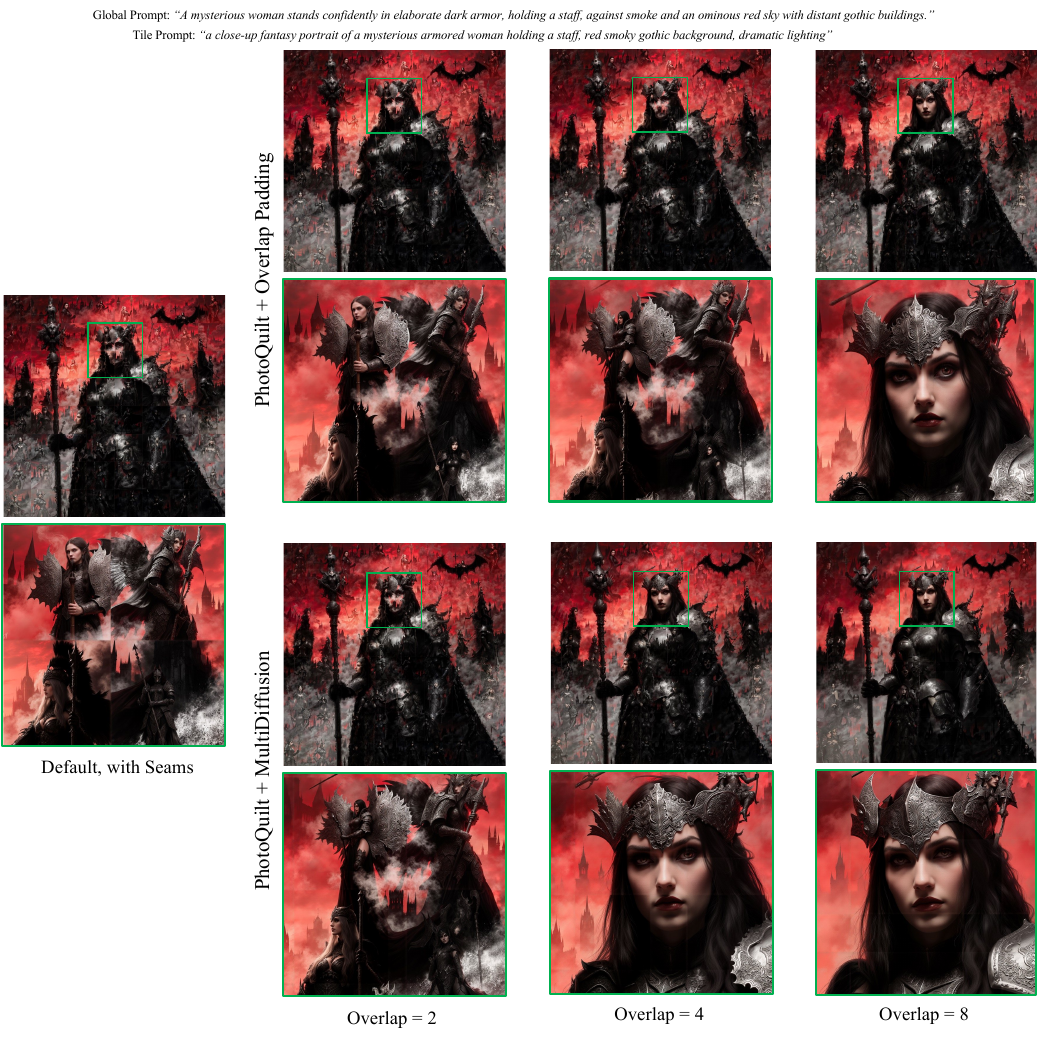}
  \caption{Qualitative comparison between different values of overlap padding across two approaches handling seams.}
  \label{fig:overlap-padding}
\end{figure*}

\begin{table*}[t]
\centering
\caption{Grid density and mosaic resolution. The default $8\times8$ configuration
is reported in the main paper. Global structure at $64\times64$. Best per
configuration in \textbf{bold}, second \underline{underlined}}.
\label{tab:grid_density}
\resizebox{\textwidth}{!}{
\scriptsize
\begin{tabular}{l cccccc cccccc c}
\toprule
\multirow{2}{*}{\textbf{Method}} & \multicolumn{6}{c}{\textbf{Global Structure ($64\times64$)}} & \multicolumn{6}{c}{\textbf{Local Tiles}} & \multirow{2}{*}{\textbf{DSR} $\uparrow$} \\
\cmidrule(lr){2-7} \cmidrule(lr){8-13}
& PSNR $\uparrow$ & SSIM $\uparrow$ & LPIPS $\downarrow$ & HPS $\uparrow$ & IR $\uparrow$ & FID $\downarrow$ & CLIP $\uparrow$ & BLIP $\uparrow$ & IQA $\uparrow$ & HPS $\uparrow$ & IR $\uparrow$ & FID $\downarrow$ & \\
\midrule
\multicolumn{14}{l}{\textit{$6\times6$ grid, $4608^2$ mosaic, $768^2$ tiles}} \\
\midrule
Phomosaic & 23.60\sd{1.60} & 0.73\sd{0.03} & 0.083\sd{0.034} & 0.225\sd{0.036} & -0.12\sd{1.04} & \underline{234.95} & 0.330\sd{0.031} & \underline{0.83}\sd{0.20} & 0.80\sd{0.12} & 0.184\sd{0.031} & -0.83\sd{0.62} & 337.19\sd{27.3} & 0.369 \\
PhotoQuilt (SD2.1) & \underline{24.36}\sd{1.61} & \underline{0.80}\sd{0.05} & \underline{0.080}\sd{0.044} & \underline{0.229}\sd{0.032} & \underline{-0.08}\sd{1.03} & \textbf{233.67} & \underline{0.332}\sd{0.032} & 0.82\sd{0.19} & \textbf{0.85}\sd{0.10} & \underline{0.192}\sd{0.029} & \underline{-0.74}\sd{0.52} & \underline{319.04}\sd{31.0} & \underline{0.386} \\
PhotoQuilt (FLUX.1) & \textbf{28.27}\sd{1.81} & \textbf{0.88}\sd{0.04} & \textbf{0.048}\sd{0.038} & \textbf{0.239}\sd{0.037} & \textbf{0.19}\sd{0.95} & 236.24 & \textbf{0.350}\sd{0.029} & \textbf{0.94}\sd{0.07} & \underline{0.82}\sd{0.12} & \textbf{0.231}\sd{0.032} & \textbf{0.13}\sd{0.65} & \textbf{317.97}\sd{30.6} & \textbf{0.540} \\
\midrule
\multicolumn{14}{l}{\textit{$12\times12$ grid, $6144^2$ mosaic, $512^2$ tiles}} \\
\midrule
Phomosaic & 23.04\sd{1.40} & 0.76\sd{0.03} & \underline{0.066}\sd{0.029} & \textbf{0.238}\sd{0.035} & \underline{-0.03}\sd{0.87} & \textbf{235.83} & \underline{0.331}\sd{0.030} & \underline{0.81}\sd{0.16} & 0.77\sd{0.12} & 0.191\sd{0.029} & -0.75\sd{0.59} & \underline{275.78}\sd{24.1} & 0.389 \\
PhotoQuilt (SD2.1) & \underline{24.46}\sd{1.43} & \underline{0.80}\sd{0.06} & 0.094\sd{0.050} & \underline{0.234}\sd{0.034} & \textbf{0.01}\sd{0.98} & \underline{241.87} & \underline{0.331}\sd{0.028} & 0.80\sd{0.16} & \underline{0.85}\sd{0.09} & \underline{0.195}\sd{0.027} & \underline{-0.67}\sd{0.50} & \textbf{263.38}\sd{26.8} & \underline{0.405} \\
PhotoQuilt (FLUX.1) & \textbf{28.07}\sd{1.67} & \textbf{0.87}\sd{0.05} & \textbf{0.056}\sd{0.052} & 0.223\sd{0.038} & -0.09\sd{1.07} & 248.41 & \textbf{0.364}\sd{0.027} & \textbf{0.98}\sd{0.03} & \textbf{0.86}\sd{0.10} & \textbf{0.254}\sd{0.035} & \textbf{0.62}\sd{0.61} & 278.14\sd{29.0} & \textbf{0.552} \\
\bottomrule
\end{tabular}
}
\end{table*}

\section{Overlap Padding for Seams}
\label{sec:overlap_pad}
Seams are the visible lines between tiles in a photomosaic. They are expected, and often preferred, since they are a signature of the mosaic form and mark the tiles as independent images. However, someone may prefer them removed, so that tiles blend into one another while keeping their independent character. PhotoQuilt supports this through an overlap padding feature. When it is enabled, the tokens of a tile additionally and randomly attend to a band of tokens on a customizable border around that tile. Each tile still follows its own trajectory and still produces its own output (queries scope is intact, only key-value scope is randomly expanded); the padding only lets a tile see its neighbours while it does so. Fig.~\ref{fig:overlap-padding} shows a prompt at several padding widths. At the default padding of 0 the seams are visible. At 2 or 4, they are removed. Increasing the padding further causes tiles to lose their independent appearance and merge together. As seen in the first example at padding 8, four tiles undesirably merge to form a single face.

This exposes tile coupling as a controllable parameter, and it is instructive to compare it with the coupling used by tiled high-resolution methods. MultiDiffusion and its successors also generate on overlapping regions, but they reconcile them by averaging the predictions of overlapping windows at every denoising step. The two mechanisms share a goal and differ in where information flows. Attention lets a tile observe its neighbours and adapt its own content to them; averaging replaces the content of a region with the mean of two independent predictions. Averaging two plausible latents does not in general yield a plausible latent, and the resulting blur and scene blending is a known limitation of this family. Fig.~\ref{fig:overlap-padding} includes this variant for comparison: at matched overlap width, averaging moves the seams instead of removing them, and degrades the tiles themselves, whereas padding removes the seams while each tile remains a single coherent generation. The cost also differs, since averaging requires every overlap region to be denoised once per window covering it and prevents parallelization and distribution, while padding widens the attention window but denoises each tile once. We keep padding 0 as the default, as a photomosaic is conventionally a grid of discrete images.

\begin{figure*}[pt!]
  \centering
  \includegraphics[width=0.95\textwidth]{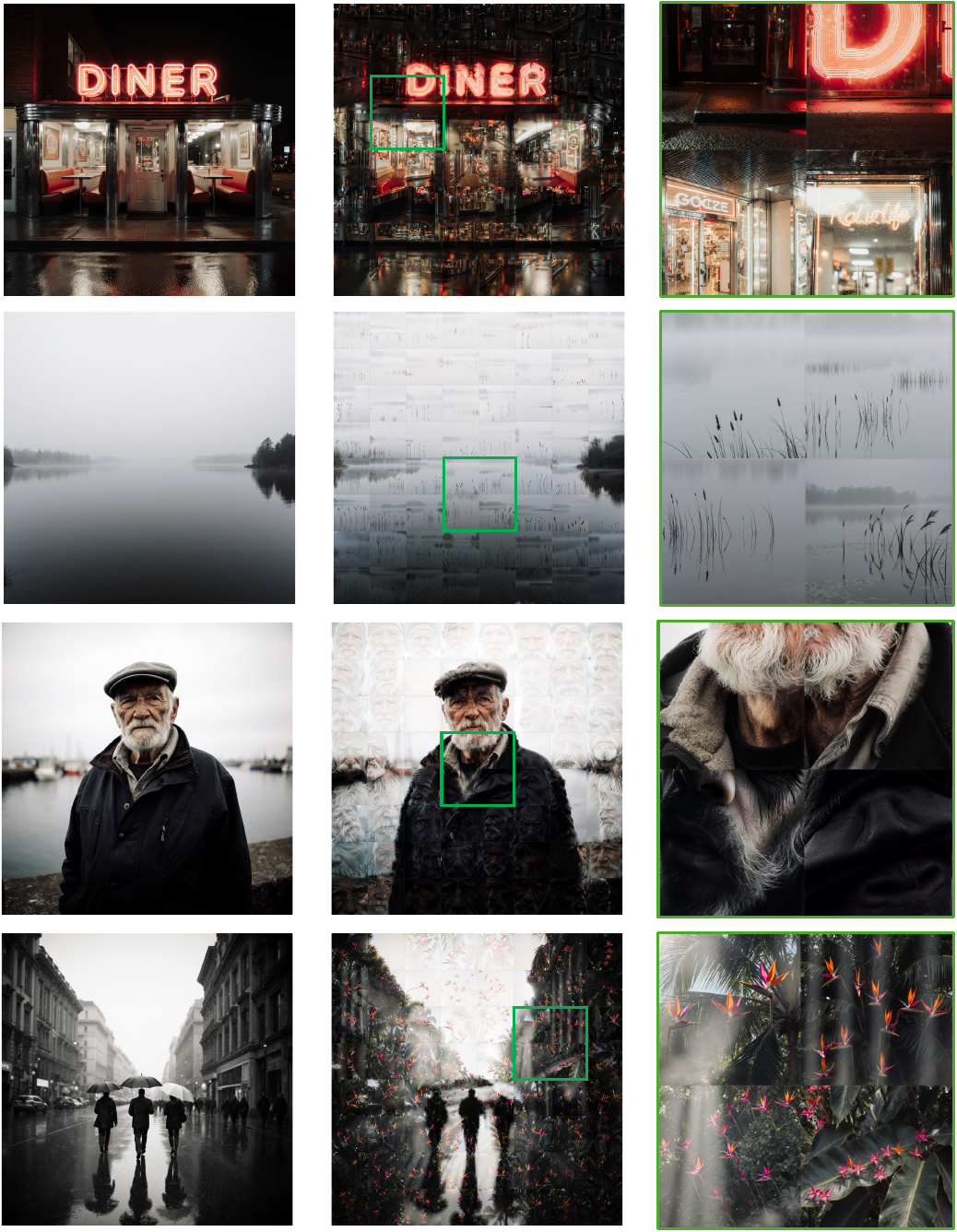}
  \caption{Failure cases, all at the default configuration. Left to right: the $768\times768$ base image, the $6144\times6144$ photomosaic, and a crop of four adjacent tiles. Best viewed zoomed in.}
  \label{fig:failures}
\end{figure*}

\begin{figure*}[pt!]
  \centering
  \includegraphics[width=0.95\textwidth]{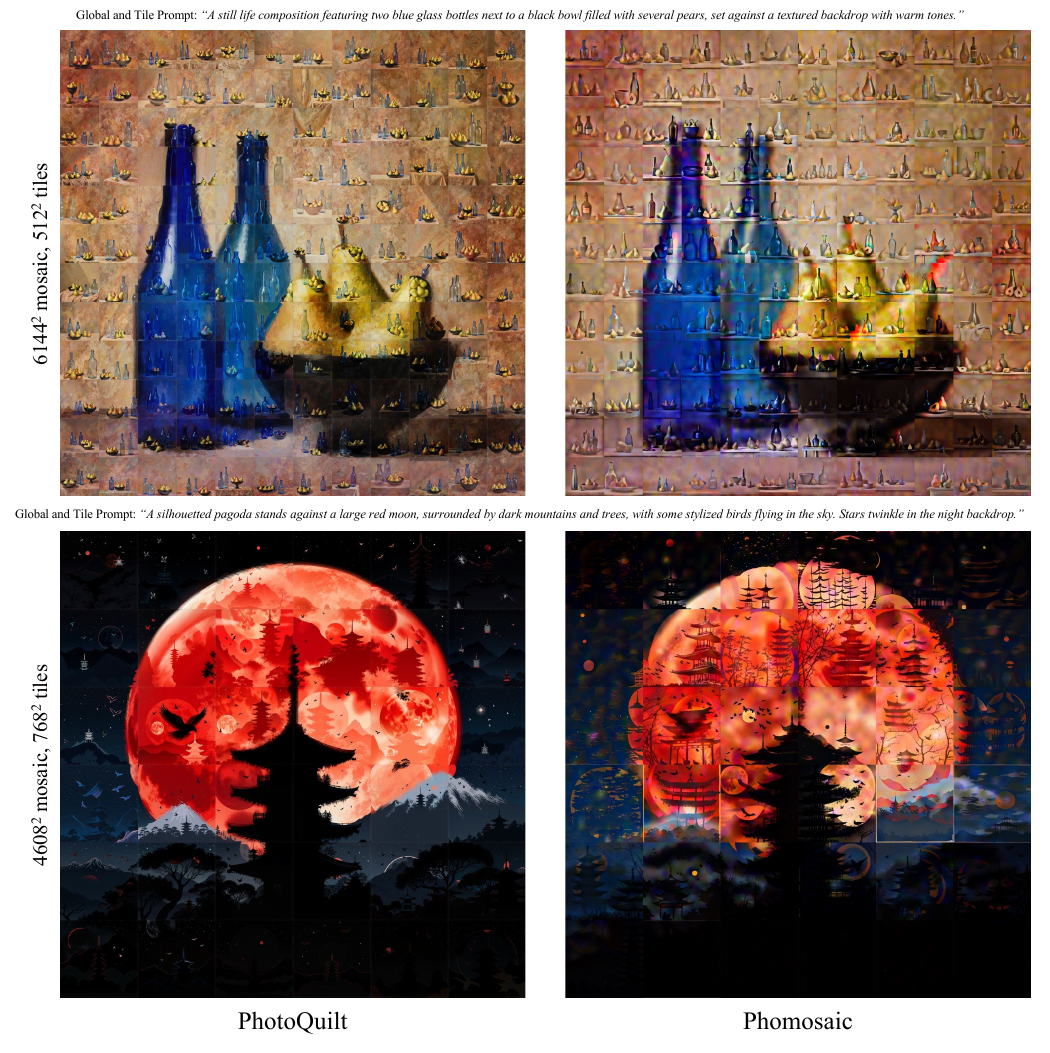}
  \caption{Grid density and mosaic resolution. PhotoQuilt preserves the global image at both settings while keeping tiles distinct. Best viewed zoomed in.}
  \label{fig:grid_density}
\end{figure*}

\section{Failure Cases}

We report four failure cases, all produced with the default configuration rather than adversarial settings. Fig.~\ref{fig:failures} shows each case as the base image, the resulting photomosaic, and a crop of four adjacent tiles.

\textbf{Over-constrained tiles.} The shared renoised latent constrains each tile in proportion to the local contrast of the base image, and where that contrast is extreme the tile is left almost no freedom. The first row uses a night storefront whose sign carries large high-contrast lettering. The target is reconstructed faithfully, but the tiles covering the strokes are dictated by them: rather than self-contained images they become renderings of letter fragments. Criterion (i) holds while criterion (ii) fails.

\textbf{Under-constrained tiles.} The same relation fails at the opposite extreme. The second row uses a lake under dense fog, with almost the entire frame filled by featureless mist. With little low-frequency structure to follow, tiles diverge freely and fill the region with foliage and branches, so the calmest part of the target becomes the busiest part of the mosaic and criterion (i) fails where the target is simplest.

\textbf{Absent cross-tile agreement.} Tiles receive no information about their neighbours, so any structure spanning a seam is completed independently on both sides. In the third row the subject's head crosses a seam intersection and is reconstructed with duplicated neck contours and inconsistent features across the boundary. The global image still reads as a portrait, since the low frequencies place the head correctly; the failure is at the intermediate scale.

\textbf{Appearance follows the tile condition.} The shared latent fixes spatial layout but not appearance, which each tile regenerates. In the fourth row the global condition is a monochrome rainy street and the tile condition a saturated floral close-up. The composition survives, with the figures and the street still in place, but colour and mood follow the tile condition rather than the global one.

\begin{figure*}[pt!]
  \centering
  \includegraphics[width=0.95\textwidth]{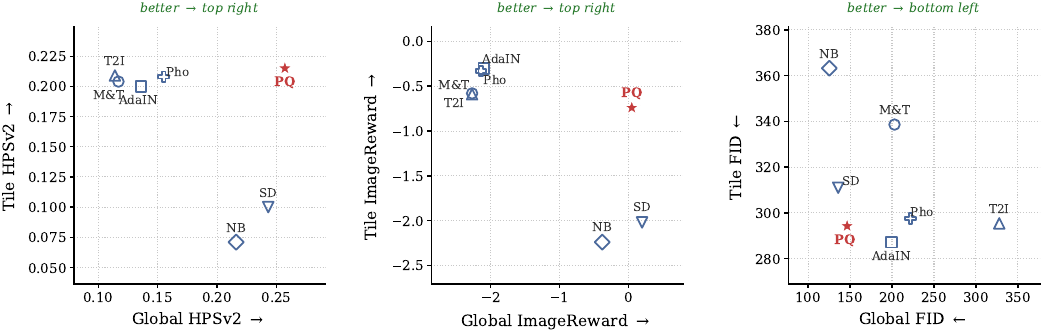}
  \caption{Dual-scale behaviour of the three metrics computed at both scales. Each panel plots the global-scale value against the tile-scale value; the preferred corner is indicated above each panel. Baselines separate into two groups: methods that reach high global fidelity (NoiseBlend, StreamDiff) collapse at the tile scale, while methods with strong tile quality (Match~\&~Tone, Color T2I-Adapter, AdaIN, Phomosaic) fail at the global scale. PhotoQuilt (PQ) is the only method that avoids both extremes. All methods use SD2.1; values from Table. 1 of the main manuscript.}
  \label{fig:dsr_scatter}
\end{figure*}

\section{Grid Density and Mosaic Resolution}

Our main evaluation uses a single configuration: a $6144\times6144$ mosaic partitioned into an $8\times8$ grid of $768\times768$ tiles. To test whether the method depends on that choice, we evaluate two further configurations. The first holds the canvas fixed and reduces the tile size to $512\times512$, giving a denser $12\times12$ grid of $144$ tiles. The second holds the tile size fixed at $768\times768$ and reduces the canvas to $4608\times4608$, giving a sparser $6\times6$ grid of $36$ tiles. Together these vary tile count by a factor of four and mosaic resolution by a third, in both directions from the default. We compare against Phomosaic, the prior generative photomosaic method, under identical settings.

Results are given in Table~\ref{tab:grid_density} and qualitative examples in Fig.~\ref{fig:grid_density}. PhotoQuilt on FLUX.1 gives the best dual-scale reward and its SD2.1 variant is the second best at every configuration, and its global and local scores move only slightly across a fourfold change in tile count, confirming that the bootstrapped formulation is not tied to a particular grid. 

\section{Dual-Scale Reward}
\label{sec:dsr}

Our two metric families measure the two criteria of a photomosaic separately, and no single number reports whether a method satisfies both. A method can therefore appear competitive by excelling at one scale while failing at the other, as Fig.~\ref{fig:dsr_scatter} shows: every baseline sits at one extreme or the other, and the two groups separate along opposite axes, while PhotoQuilt balances the dual-scale evaluation across three metrics.

We summarise the two scales with a single score built on Image Reward, one metric we compute at both scales and a learned model of human preference rather than a hand-designed similarity. Let $r_g$ and $r_t$ be the mean Image Reward of the downsampled mosaic and of its tiles, as reported in the corresponding columns of our tables. Image Reward is unbounded and takes negative values, so we map each to $(0,1)$ with the logistic $\sigma(x) = 1/(1+e^{-x})$ and take the harmonic mean,
\begin{equation}
\mathrm{DSR} \;=\; \frac{2\,\sigma(r_g)\,\sigma(r_t)}{\sigma(r_g)+\sigma(r_t)}.
\label{eq:dsr}
\end{equation}
Dual-scale Reward (DSR) is a deterministic function of the two reported values, so every entry can be recomputed directly from the table. The ordering induced by Eq.~\ref{eq:dsr} is not an artefact of the choice of aggregator. The harmonic mean of sigmoids is deliberately dominated by the weaker term, so a method with one catastrophic scale cannot score well regardless of the other. Replacing the harmonic mean by the geometric mean leaves the ranking in tables unchanged, and replacing it by the minimum of the two terms changes only the order of the two lowest-scoring methods, which lie within $0.001$ of one another under all three formulations.

\begin{table}[t]
\centering
\caption{Human perceptual study with $20$ participants. Ratings are on a five-point scale; preference is a forced choice over $180$ trials, nine per participant. All methods use SD2.1.}
\label{tab:user_study}
\resizebox{\columnwidth}{!}{
\scriptsize
\begin{tabular}{l cc ccc}
\toprule
\multirow{2}{*}{\textbf{Method}} & \multicolumn{2}{c}{\textbf{Preference}} & \multicolumn{3}{c}{\textbf{Rating}} \\
\cmidrule(lr){2-3} \cmidrule(lr){4-6}
& \% $\uparrow$ & Picks & Global $\uparrow$ & Tile $\uparrow$ & Overall $\uparrow$ \\
\midrule
Match \& Tone & 2.8 & 5/180 & 1.16 & \textbf{4.51} & 1.58 \\
AdaIN         & 1.1 & 2/180 & 1.75 & 3.65 & 1.75 \\
StreamDiff    & \underline{17.8} & 32/180 & 3.95 & 2.10 & 2.83 \\
Phomosaic     & 13.9 & 25/180 & \underline{3.97} & 2.83 & \underline{3.31} \\
\midrule
PhotoQuilt (SD2.1) & \textbf{64.4} & 116/180 & \textbf{4.33} & \underline{4.28} & \textbf{4.25} \\
\bottomrule
\end{tabular}
}
\end{table}

\section{Human Perceptual Study}
\label{sec:user_study}

The DSR metric described in Sec.~\ref{sec:dsr} serves as an indicator of the photomosaic effect’s quality. In addition, we also conducted a human perceptual study to further strengthen our evaluation. The two lowest-scoring baselines were excluded to keep sessions under 20 minutes. Participants completed four tasks. The first showed a downsampled and zoomed-out mosaic and asked how well it reads as a global scene. The second showed tiles at native resolution and asked for two ratings, tile quality and tile diversity. The third showed the full mosaic together with a tile crop and asked for a single photomosaic score, judging both scales at once. The fourth showed the outputs of all five methods for the same prompt and asked for the preferred result. Ratings use a five-point scale and the preference task is a forced choice. Twenty participants each completed nine preference trials, giving $180$ trials in total. Methods were unlabelled and presentation order was randomised. Participants were recruited by open invitation and were not screened for expertise.

Results are given in Table~\ref{tab:user_study}. The study reproduces the trade-off our metrics report. Match \& Tone obtains the highest tile rating in the study ($4.51$) together with the lowest global rating ($1.16$), while StreamDiff obtains the lowest tile rating ($2.10$) with a strong global rating ($3.95$). These are the two extremes of Fig.~\ref{fig:dsr_scatter}, now measured by human perception rather than by proxy.

PhotoQuilt is not the best method in the tile scale. Tile rating of $4.28$ is below Match \& Tone's, and global rating of $4.33$ exceeds the next method by $0.36$. It is selected in $64.4\%$ of forced-choice trials, more than three times the next method. Also, the DSR ordering agrees with the preference ordering on the two leading methods and places the remaining three within a narrow band in Table 1.


\end{document}